# Model-Based Diagnosis using Structured System Descriptions

**Adnan Darwiche**                                      DARWICHE@AUB.EDU.LB
*Department of Mathematics*
*American University of Beirut*
*PO Box 11-236*
*Beirut, Lebanon*

## Abstract

This paper presents a comprehensive approach for model-based diagnosis which includes proposals for characterizing and computing preferred diagnoses, assuming that the system description is augmented with a system structure (a directed graph explicating the interconnections between system components). Specifically, we first introduce the notion of a *consequence,* which is a syntactically unconstrained propositional sentence that characterizes all consistency-based diagnoses and show that standard characterizations of diagnoses, such as minimal conflicts, correspond to syntactic variations on a consequence. Second, we propose a new syntactic variation on the consequence known as *negation normal form (NNF)* and discuss its merits compared to standard variations. Third, we introduce a basic algorithm for computing consequences in NNF given a structured system description. We show that if the system structure does not contain cycles, then there is always a linear–size consequence in NNF which can be computed in linear time. For arbitrary system structures, we show a precise connection between the complexity of computing consequences and the topology of the underlying system structure. Finally, we present an algorithm that enumerates the preferred diagnoses characterized by a consequence. The algorithm is shown to take linear time in the size of the consequence if the preference criterion satisfies some general conditions.

## 1. Introduction

This paper presents a comprehensive approach for characterizing and computing preferred diagnoses when the system description is augmented with a system structure (Darwiche, 1995). A system structure is a directed acyclic graph explicating the interconnections between system components. Adding a system structure to a classical system description (de Kleer, Mackworth, & Reiter, 1992) leads to what we call a *structured system description,* examples of which are shown in Figures 1 and 2.

The most common approach for characterizing (and computing) diagnoses has been the use of conflicts and their derivatives such as kernel diagnoses (de Kleer & Williams, 1987; Reiter, 1987; de Kleer et al., 1992). Moreover, the most common method for computing these characterizations has been the use of Assumption-Based Truth Maintenance Systems (ATMSs) (de Kleer, 1986; Reiter & de Kleer, 1987; Forbus & de Kleer, 1993). We will first explain the difficulties with such an approach and then describe the elements of our approach that address these difficulties.

The major problem with standard characterizations of diagnoses is that they tend to be exponential in size, which is largely due to their syntactic nature. Specifically, standard characterizations of diagnoses correspond to the prime implicants/implicates of some propositional sentence. And the number of such implicants/implicates tend to be exponential even for expressions that correspond to simple diagnosis problems. Computationally, this problem manifests as an ATMS label that has an exponential number of environments (Forbus & de Kleer, 1993).

This difficulty has led to a body of research on "focusing" the ATMS, which attempts to control the size of ATMS labels (Provan, 1996; de Kleer, 1992; Forbus & de Kleer, 1988; Dressler & Farquar, 1990; Collins & DeCoste, 1991). Focusing is based on the following intuition. A label characterizes all diagnoses of a given problem, but one is rarely interested in all diagnoses; therefore, one rarely needs a complete label. Most often, one is interested in diagnoses that satisfy some preference criterion, such as most-probable diagnoses. Therefore, one can use such a criterion to compute "focused" labels that are of reasonable size, yet are good enough to characterize the diagnoses of interest. However, although a standard framework exists for computing ATMS labels (Forbus & de Kleer, 1993), no such framework seems to exist for focusing.

Another issue with standard frameworks for computing diagnoses (based on minimizing propositional sentences) is that their computational complexity is not formally tied to properties of system descriptions that are easily accessible to engineers who would be constructing these descriptions. An example of such a property is the topology of a system structure (component interconnectivity). Providing computational complexity guarantees in terms of such properties can be extremely useful in practice, as our experience has shown. In real-world applications, one may have a choice of what failures to include in the scope of a diagnostic system, and therefore a choice of what aspects of a complex system to include in a system description. In such situations, it is very important to be able to assess the effectiveness of diagnosis algorithms by an intuitive examination of the resulting system description, such as examining the topology of a structured system description. The framework we shall develop in this paper addresses this particular point and has proven very useful in helping us engineer system descriptions on which our diagnostic algorithms are guaranteed to be effective.

The approach we present in this paper is based on three main ideas, which address the problems mentioned above:

1. **Characterizing diagnoses using negation normal forms:** We propose the notion of a consequence which is a syntactically unconstrained propositional sentence that characterizes all consistency-based diagnoses (Darwiche, 1995, 1997). We show that standard characterizations of diagnoses correspond to syntactic restrictions on a consequence. Specifically, minimal conflicts correspond to the prime implicates of a consequence and kernel diagnoses correspond to its prime implicants. We adopt a less restrictive syntax of consequences known as *negation normal form (NNF)* of which prime implicants/implicates are a special case (Barwise, 1977). Although we do not guarantee that our NNF representation of consequences is the most compact, we do offer some guarantees on this representation that cannot be offered with respect to standard representations.

2. **Utilizing system structure in computing consequences:** We introduce a basic algorithm for computing consequences in NNF, the complexity of which is determined by the topology of the system structure. We show that for tree system structures (those containing no undirected cycles), there is always a consequence in NNF that is linear in the number of nodes and arcs in the structure.[1] Moreover, a standard characterization of diagnoses using minimal conflicts can be exponential in size for some of these system structures. For arbitrary system structures, we provide a precise relationship between the system structure, the size of a consequence, the time to compute it, and, hence, the difficulty of a diagnosis problem.

3. **A mechanism for computing minimal diagnoses:**[2] We show that if a consequence is in *decomposable* negation normal form, then one can extract the minimal diagnoses it characterizes in time linear in its size, as long as the minimality criterion satisfies some general conditions. The algorithm we propose for computing consequences is guaranteed to generate consequences in decomposable negation normal form. Moreover, the conditions on a minimality criterion do admit the common criterion of minimum cardinality.

Therefore, we are providing a paradigm for diagnostic reasoning with system structures, consequences, and minimality criteria as the key components. By using this paradigm, one is guaranteed some complexity results that are determined by the topology of the system structure. As we shall see, this approach is similar to the network-paradigm in the probabilistic and constraint-satisfaction literature where system structure is the key aspect that decides the difficulty of a reasoning problem.

The literature contains a number of other proposals for importing this structure-based theme into model-based diagnosis (Dechter & Dechter, 1996; Geffner & Pearl, 1987). Although these approaches appeal to similar underlying principles and lead to similar complexity results, some key differences exist between our approach and the previous ones. First, our formulation is based on symbolic logic, which is the tradition in model-based diagnosis, while the previous proposals have been based on constraints among multivalued variables. Second, the complexity of our algorithm depends not only on the system structure but also on the system observation. The stronger the system observation is, the better the complexity of our algorithm, leading to linear complexity in the extreme case (independently of the system structure). Finally, we separate the computation of minimal diagnoses into two phases: the characterization of all diagnoses using a consequence and then the extraction of minimal diagnoses from the consequence. This separation has a number of implications which are discussed in detail later in the paper.

This paper is structured as follows. We introduce the notion of a consequence in Section 2, proving that it characterizes all consistency-based diagnoses, and showing its relation to some standard notions in the literature on model-based diagnosis. We then introduce three key theorems for constructing consequences in Section 3 and formalize the role of system structure in determining the complexity of computing consequences. In Section 4, we turn to an algorithm for computing consequences when the system description is augmented with a system structure. The computational complexity of the presented algorithm is discussed at length in Section 5. We then provide an algorithm in Section 6 for extracting minimal diagnoses from a consequence. We finally close in Section 7 with some concluding remarks. Proofs of all theorems and lemmas are delegated to Appendix F.

---

1. Note, however, that the size of a consequence and the time to compute it is exponential in the size of families (each node and its parents in the device structure represent a family). In structure–based reasoning, it is typically assumed that the size of a family is small enough to be treated like a constant.

2. In this paper, we use "minimal" and "preferred" interchangeably when referring to diagnoses.

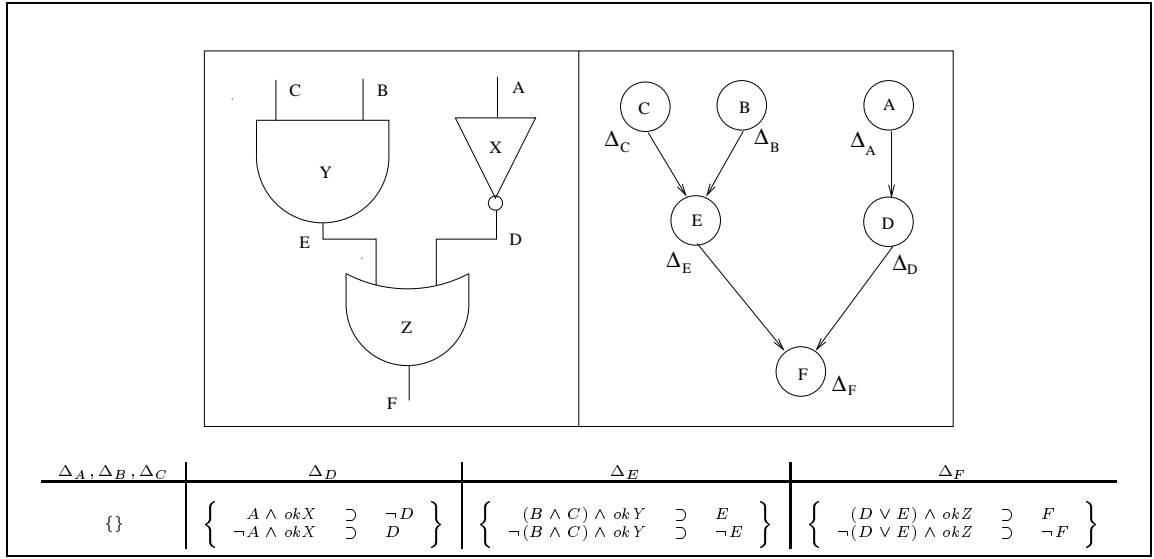

| $\Delta_A, \Delta_B, \Delta_C$ | $\Delta_D$ | $\Delta_E$ | $\Delta_F$ |
|---|---|---|---|
| {} | $\left\{ \begin{array}{l} A \wedge okX \supset \neg D \\ \neg A \wedge okX \supset D \end{array} \right\}$ | $\left\{ \begin{array}{l} (B \wedge C) \wedge okY \supset E \\ \neg(B \wedge C) \wedge okY \supset \neg E \end{array} \right\}$ | $\left\{ \begin{array}{l} (D \vee E) \wedge okZ \supset F \\ \neg(D \vee E) \wedge okZ \supset \neg F \end{array} \right\}$ |

Figure 1: A structured system description (SSD) of a digital circuit. An SSD has two parts: a directed acyclic graph and a set of component descriptions, each associated with a node in the graph. The formal definition of an SSD is given in Section 3.3.

## 2. Characterizing Diagnoses

We start in this section with a review of model-based diagnosis and then lead into the notion of a consequence for characterizing diagnoses. We show how standard characterizations of diagnoses, such as minimal conflicts, can be viewed as syntactic variations on the consequence and then introduce a new syntactic variation known as negation normal form. We also discuss our reasons for adopting this non-standard form for characterizing diagnoses.

### 2.1 Model-Based Diagnosis

In model-based diagnosis, we use the term *system description* to denote a system model (de Kleer et al., 1992; Reiter, 1987). Traditionally, a system description consists of a set of logical sentences $\Delta$ called a *database* and a set of distinguished symbols $\mathbf{A} = \{okX, okY, \ldots\}$ called *assumables*. Assumables represent the health of components and are initially assumed to be true. For example, in Figure 2, the assumables are $okX$, $okY$ and the database $\Delta$ contains the four propositional sentences shown in the figure.

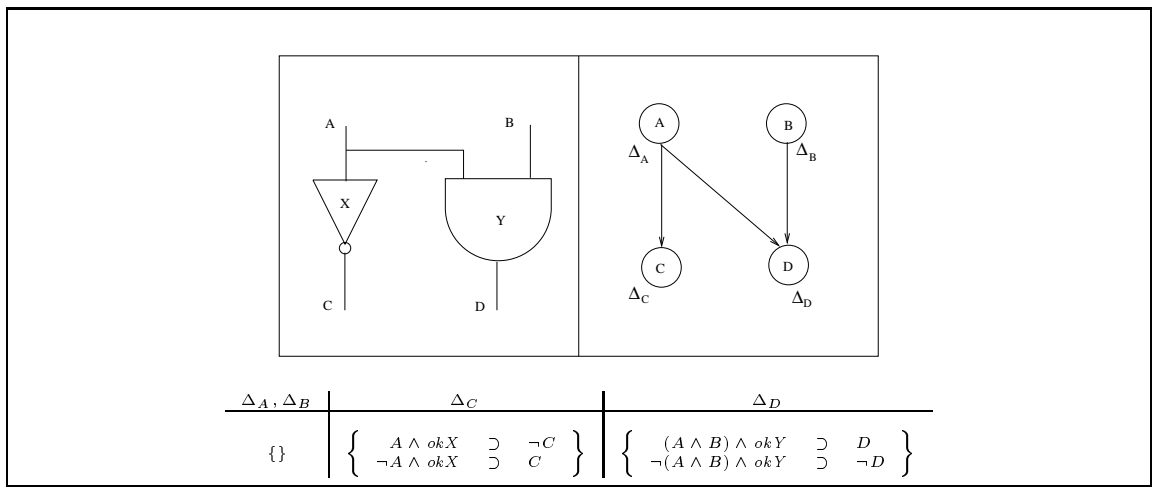

Figure 2: A structured system description of a digital circuit.

A diagnosis problem emerges when assumables can no longer be justified. Specifically, given some sentence $\phi$ that represents an observed system behavior, the system is considered faulty if $\phi$ is inconsistent with $\Delta \cup \mathbf{A}$. In this case, one needs to relax some of the assumables (that is, replace instances of $ok$. with instances of $\neg ok$.) in order to restore consistency. A particular relaxation of these assumables is called a *diagnosis* as long as it is consistent with the system description and observation. The cardinality of a diagnosis is the number of faults contained in the diagnosis. In Figure 2, a system observation $C \wedge D$ would indicate a failure. Moreover, there are three diagnoses in this case: $okX \wedge \neg okY$, $\neg okX \wedge okY$, $\neg okX \wedge \neg okY$, with cardinalities 1, 1 and 2, respectively.

We have the following formal definition of a system description, which we adopt in the rest of this paper. The definition is a variation on the standard one provided in (Reiter, 1987) and is preceded by some preliminary definitions.

**Definition 1** *Let* $\mathbf{S}$ *be a set of atomic propositions (atoms). An* <u>$\mathbf{S}$–literal</u> *is a literal whose atom is in* $\mathbf{S}$. *An* <u>$\mathbf{S}$–sentence</u> *is a propositional sentence in which each literal is an* $\mathbf{S}$–literal. *An* <u>$\mathbf{S}$–instantiation</u> *is a conjunction of* $\mathbf{S}$–literals, *one literal for each atom in* $\mathbf{S}$.

For example, if $\mathbf{S} = \{A, B, C\}$, then there are eight $\mathbf{S}$–instantiations: $A \wedge B \wedge C$, $A \wedge B \wedge \neg C$, ..., $\neg A \wedge \neg B \wedge \neg C$. Moreover, $A \wedge \neg B \supset C$ and $A \supset C$ are $\mathbf{S}$–sentences, but $A \supset D$ is not.

**Definition 2 (System Description)** *A system description is a triple* $(\mathbf{P}, \mathbf{A}, \Delta)$, *where* $\mathbf{P}$ *and* $\mathbf{A}$ *are sets of atomic propositions such that* $\mathbf{P} \cap \mathbf{A} = \emptyset$, *and* $\Delta$ *is a set of propositional sentences constructed from atoms in* $\mathbf{P}$ *and* $\mathbf{A}$. *Here,* $\mathbf{P}$ *is called the set of non-assumables;* $\mathbf{A}$ *is called the set of assumables;* $\Delta$ *is called a database. It is required that* $\Delta$ *be consistent with every* $\mathbf{A}$–instantiation.

Given the notion of a system description, we can define the other two key terms in model-based diagnosis (de Kleer et al., 1992):

**Definition 3 (Observation)** *Given a system description* $(\mathbf{P}, \mathbf{A}, \Delta)$, *a system observation is a consistent conjunction of* $\mathbf{P}$*–literals.*

**Definition 4 (Diagnosis)** *Given a system description* $(\mathbf{P}, \mathbf{A}, \Delta)$ *and a system observation* $\phi$, *a diagnosis is an* $\mathbf{A}$*–instantiation that is consistent with* $\Delta \cup \{\phi\}$.

It is worth mentioning here that the condition we imposed on database $\Delta$ in Definition 2 is equivalent to saying that there are no diagnoses for the system observation *true*. That is, if we have no observation about the system, then we cannot conclude anything about the health of its components.[3]

## 2.2 The Consequence

An ultimate goal of diagnostic reasoning is to compute the minimal diagnoses (according to some criterion) for a given system description $(\mathbf{P}, \mathbf{A}, \Delta)$ and observation $\phi$. Standard approaches to model-based diagnoses do this in conceptually two steps. First, they characterize the set of diagnoses using conflicts and then extract minimal diagnoses from this characterization.

We will follow the same approach except that we will not use conflicts to characterize the set of diagnoses. Instead, we will adopt an equivalent but syntactically different characterization of diagnoses known as negation normal form (Barwise, 1977). Before we discuss this alternate characterization of diagnoses, however, we need to introduce the notion of a consequence which is very useful in putting the different characterizations of diagnoses in perspective.

The consequence of an observation is defined formally below (Darwiche, 1995, 1997):

**Definition 5 (Consequence)** *Given a system description* $(\mathbf{P}, \mathbf{A}, \Delta)$, *the consequence of system observation* $\phi$, *written* $Cons_{\mathbf{A}}^{\Delta}(\phi)$, *is a sentence satisfying the following properties:*

1. *$Cons_{\mathbf{A}}^{\Delta}(\phi)$ is an $\mathbf{A}$-sentence;*

2. *$\Delta \cup \{\phi\} \models Cons_{\mathbf{A}}^{\Delta}(\phi)$;*

3. *For any $\mathbf{A}$-sentence $\beta$, $\Delta \cup \{\phi\} \models \beta$ only if $Cons_{\mathbf{A}}^{\Delta}(\phi) \models \beta$.[4]*

That is, the consequence of observation $\phi$ is the logically strongest $\mathbf{A}$-sentence which is entailed by the system description $\Delta$ and observation $\phi$.

When clear from the context, we drop the superscript $\Delta$, the subscript $\mathbf{A}$, or both, from the notation $Cons_{\mathbf{A}}^{\Delta}$.

---

3. This condition is intuitive if we are not representing fault modes and are restricting ourselves to using the *ok.* assumables. However, if we use fault modes, such as *stuck-at-0* and *stuck-at-1*, then the condition is not as intuitive and may even appear restrictive. Specifically, we may want to constrain these two assumables such that $\neg$ *stuck-at-0* $\lor$ $\neg$ *stuck-at-1* is true. But introducing such a constraint would violate the condition we are imposing on $\Delta$ in Definition 2. We will make two points regarding this issue. First, one can represent constraints among assumables without violating the above condition; the details of this are explained in Appendix D. However, a better solution to this problem is to use multivalued variables instead of atomic propositions, which leads to a more convenient, but less standard framework, which is also discussed in Appendix D.

4. The consequence of an observation is unique up to logical equivalence.

For example, if we observe that both $C$ and $D$ are true in the circuit of Figure 2, we conclude that one of the gates must be malfunctioning: under normal conditions, $C$ being true implies that $A$ is false, which further implies that $D$ is false. Therefore, a conclusion one can make about assumables, having observed $C \wedge D$, is that $\neg okX \vee \neg okY$. Moreover, this is the strongest conclusion that can be made given the system description and observation at hand. Formally, this means that $\neg okX \vee \neg okY$ is the consequence of observation $C \wedge D$, $Cons_{\mathbf{A}}^{\Delta}(C \wedge D)$.

That a consequence characterizes all consistency-based diagnoses is shown formally below:

**Theorem 1 (Characterization)** *Given a system description* $(\mathbf{P}, \mathbf{A}, \Delta)$*, an* $\mathbf{A}$*-instantiation* $\alpha$ *is a diagnosis for system observation* $\phi$ *according to Definition 4 iff* $\alpha \models Cons_{\mathbf{A}}^{\Delta}(\phi)$.

For example, the consequence $\neg okX \vee \neg okY$ characterizes three diagnoses: $\neg okX \wedge \neg okY$, $okX \wedge \neg okY$ and $\neg okX \wedge okY$.

Standard characterizations of diagnoses can be viewed as syntactic restrictions on a consequence. Consider the following theorem first, which is a corollary of the results reported in (de Kleer et al., 1992):

**Theorem 2** *Given a system description* $(\mathbf{P}, \mathbf{A}, \Delta)$ *and observation* $\phi$*, we have the following:*

- *A* <u>partial diagnosis</u> *is any* $\mathbf{A}$*-sentence which is an implicant of* $Cons_{\mathbf{A}}^{\Delta}(\phi)$.

- *A* <u>kernel diagnosis</u> *is any* $\mathbf{A}$*-sentence which is a prime implicant of* $Cons_{\mathbf{A}}^{\Delta}(\phi)$.

- *A* <u>conflict</u> *is any* $\mathbf{A}$*-sentence which is an implicate of* $Cons_{\mathbf{A}}^{\Delta}(\phi)$.

- *A* <u>minimal conflict</u> *is any* $\mathbf{A}$*-sentence which is a prime implicate of* $Cons_{\mathbf{A}}^{\Delta}(\phi)$.

Prime implicates and implicants are standard notions but we include their definitions here for completeness:

**Definition 6** *An* <u>implicant</u> $\beta$ *of a sentence* $\alpha$ *is a satisfiable conjunction of literals which entails* $\alpha$*. We say that* $\beta$ *is a* <u>prime implicant</u> *of* $\alpha$ *if no subset of its literals satisfies this condition. An* <u>implicate</u> $\beta$ *of a sentence* $\alpha$ *is a non-valid disjunction of literals which is entailed by* $\alpha$*. We say that* $\beta$ *is a* <u>prime implicate</u> *of* $\alpha$ *if no subset of its literals satisfies this condition.*

Note that each sentence is equivalent to the conjunction of its prime implicates. Therefore, the conjunction of minimal conflicts is nothing but a syntactic form of the consequence. Similarly, each sentence is equivalent to the disjunction of its prime implicants. Therefore, the disjunction of kernel diagnoses is also a syntactic form of the consequence.[5]

---

5. The notion of a consequence as we define it in this paper seems to correspond to notions that have appeared previously in the diagnostic literature, but these notions did not prove to be computationally influential. For example, Saraswat et al. define in (Saraswat, de Kleer, & Raiman, 1990) a *maximally abstract diagnosis*, which corresponds to our notion of a consequence. They also prove that a maximally abstract diagnosis characterizes the set of consistency-based diagnoses in the sense of Theorem 1 above. Similarly, Ayeb el al. define in (el Ayeb, Marquis, & Rusinowitch, 1993) a *deductive diagnosis*, which is closely related to a consequence — the logically strongest deductive diagnosis is the consequence. In both cases, however, the proposed notions are not utilized computationally as we shall utilize consequences in the rest of this paper.

### 2.3 Negation Normal Form

The minimal-conflict representation of a consequence is a minimized conjunctive normal form (CNF) and the kernel-diagnoses representation is a minimized disjunctive normal form (DNF). Both of these standard forms, however, are specials cases of negation normal form (NNF) (Barwise, 1977), which we adopt in this paper:

**Definition 7 (Negation Normal Form)** *A sentence $\alpha$ is in negation normal form (NNF) if and only if $\alpha$ is either a literal; a disjunction $\bigvee_i \alpha_i$; or a conjunction $\bigwedge_i \alpha_i$ where each $\alpha_i$ is in negation normal form.*

That is, in an NNF, the negation operator can only be applied to atoms — it cannot be applied to compound sentences. For example, $\neg(A \wedge B)$ is not in NNF, but $\neg A \vee \neg B$ is. An NNF will typically have nested conjunctions of disjunctions and nested disjunctions of conjunctions.

The algorithm we present later will generate consequences in negation normal form. The generated consequences are not guaranteed to be the most compact, but we shall offer some guarantees on their sizes that do not hold with respect to consequences in standard forms.

The consequences we shall generate are not only in negation normal form but are also decomposable:

**Definition 8 (Decomposable)** *A sentence $\alpha$ in NNF is decomposable if and only if no atoms are shared by any conjuncts in $\alpha$.*[6]

The sentence $((\neg A \vee \neg okX) \wedge (\neg C \vee \neg okY)) \vee ((\neg okX \vee A) \wedge (\neg okY \vee C))$ is in decomposable negation normal form: the negation operator appears only next to atoms in the sentence; and there are no common atoms between the conjuncts $\neg A \vee \neg okX$ and $\neg C \vee \neg okY$, neither there are any common atoms between the conjuncts $\neg okX \vee A$ and $\neg okY \vee C$.

The decomposability property is quite strong because it allows one to decompose certain computations with respect to an NNF into smaller computations with respect to its subsentences. In Section 6, we shall see how this decomposability property will allow us to extract the minimal diagnoses characterized by a consequence by simply combining the minimal diagnoses characterized by its subsentences.[7]

Throughout the paper, we will be representing negation normal forms using directed acyclic graphs. This representation is detailed in Appendix A which also provides a number of operations for manipulating this graphical representation of negation normal forms. The operations defined in this appendix will be used in the pseudocode that we present later for generating consequences.

---

6. Every DNF is also a decomposable NNF. But this is not necessarily true for CNFs.

7. To appreciate this decomposability property, consider testing satisfiability as an example. It is easy to verify that such a test can be performed in time which is linear in the size of an NNF if the NNF is decomposable. In particular:

1. if $\alpha$ is a literal, then $\alpha$ is satisfiable;

2. if $\alpha = \alpha_1 \vee \ldots \vee \alpha_n$, then $\alpha$ is satisfiable iff some $\alpha_i$ is satisfiable;

3. if $\alpha = \alpha_1 \wedge \ldots \wedge \alpha_n$, then $\alpha$ is satisfiable iff every $\alpha_i$ is satisfiable.

Case 3 above does not hold in general since each of two sentences may be satisfiable but their conjunction may not. However, if the two sentences do not share any atoms, then their satisfiability is enough to guarantee the satisfiability of their conjunction.

## 3. Computing Consequences: Some Fundamental Theorems

Given some system observation $\phi$ and some minimality criterion, our goal is to compute all minimal diagnoses of $\phi$ according to this criterion. We will do this in two steps. First, we will compute the consequence of $\phi$ in NNF. Second, we will extract from the computed consequence all the minimal diagnoses it characterizes. The second step will be addressed in Section 6. In this and the following section, we focus on the first step.

Our strategy for computing consequences is to construct them from component consequences. Intuitively, a component consequence is the strongest conclusion one can draw about the health of a component given a particular state of its *ports* (inputs and outputs). We shall present two theorems, Decomposition and Case-Analysis, which are sufficient to construct any consequence by logically combining component consequences. We start by discussing component consequences first and then present the mentioned theorems.

### 3.1 Component Consequences

We assume that the sentences in database $\Delta$, of a system description $(\mathbf{P}, \mathbf{A}, \Delta)$, are grouped into sets, each representing the description of some component in the system. Consider the system in Figure 2 for example, which has two components whose outputs are denoted by $C$ and $D$. The database $\Delta$ for this system is viewed as the union of two component descriptions $\Delta_C$ and $\Delta_D$ where

$$\Delta_C = \left\{ \begin{array}{lcr} A \wedge okX & \supset & \neg C \\ \neg A \wedge okX & \supset & C \end{array} \right\};$$

and

$$\Delta_D = \left\{ \begin{array}{lcr} A \wedge B \wedge okY & \supset & D \\ \neg(A \wedge B) \wedge okY & \supset & \neg D \end{array} \right\}.$$

Without loss of generality, we assume that a system component has one output and we use this output to identify the component.[8]

If $\Delta_O$ is a component description, then $Cons^{\Delta_O}_{\mathbf{A}}(\phi)$ is called a *component consequence* whenever $\phi$ is an instantiation of the ports of component $O$. Following are some component consequences with respect to Figure 2:

$$\begin{array}{rcl} Cons^{\Delta_C}(A \wedge C) & \equiv & \neg okX, \\ Cons^{\Delta_C}(A \wedge \neg C) & \equiv & true, \\ Cons^{\Delta_D}(A \wedge B \wedge D) & \equiv & true, \\ Cons^{\Delta_D}(A \wedge \neg B \wedge D) & \equiv & \neg okY. \end{array}$$

Intuitively, a component consequence is a strongest conclusion that can be made about the health of a component given an observation about that component ports.

We will now provide a theorem that will be the basis for computing a component consequence in time linear in the number of clauses in the component description. But first, the following definition:

---

8. If we want to model a component with $n$ outputs, we model it as a set of $n$ components, each with one output.

**Definition 9 (Projection)** *The projection of an instantiation (clause) $\alpha$ on a set of atoms* **S**, *written $\alpha_{\mathbf{S}}$, is the conjunction (disjunction) of all* **S**–*literals in $\alpha$. If $\Delta$ is a database, then $\alpha_{\Delta}$ is the projection of $\alpha$ on the atoms appearing in $\Delta$.*

For example, the projection of $\alpha = A \vee \neg B \vee C$ on atoms $\{A, B\}$ is $A \vee \neg B$. Moreover, the projection of $\alpha$ on database $\Delta_C$ above is $A \vee C$.

**Theorem 3 (Component-Consequence)** *Let $O$ be a component with inputs* **I** *and description $\Delta_O$ (in clausal form). If $\phi$ is an instantiation of atoms* **I** $\cup \{O\}$, *then*

$$Cons_{\mathbf{A}}^{\Delta_O}(\phi) \equiv \bigwedge_{\alpha \in \Delta_O, \ \phi \models \neg\alpha_{\mathbf{P}}} \alpha_{\mathbf{A}}.$$

Consider the component description $\Delta_D$ in Figure 2, which is shown in clausal form below:

$$\Delta_D = \left\{ \begin{array}{c} \neg A \vee \neg B \vee \neg okY \vee D \\ A \vee \neg okY \vee \neg D \\ B \vee \neg okY \vee \neg D \end{array} \right\}.$$

Consider also the observation $\phi = A \wedge \neg B \wedge D$ which represents a particular state of the component ports. For the first and second clauses $\alpha$ in $\Delta_D$, we have $\phi \models \alpha_{\mathbf{P}}$. For the third clause $\alpha$, we have $\phi \models \neg\alpha_{\mathbf{P}}$. Therefore, the component consequence in this case is $\neg okY$ which is the projection of third clause in $\Delta_D$ on the assumables.

Any consequence computed by Theorem 3 is guaranteed to be in CNF. By simply converting it to DNF, we would also be converting it to decomposable NNF since each DNF is a decomposable NNF. Note that this conversion is exponential in the number of assumables appearing in the consequence. However, the number of such assumables is typically small enough to justify viewing this conversion as taking constant time.

## 3.2 Decomposition and Case-Analysis

We now provide two theorems which are sufficient for constructing any system consequence from component consequences.

**Theorem 4 (Decomposition)** *Let* $(\mathbf{P}, \mathbf{A}, \Theta \cup \Gamma)$ *be a system description and let $\phi$ be a system observation. If the atoms shared by $\Theta$ and $\Gamma$ all appear in $\phi$, then*

$$Cons_{\mathbf{A}}^{\Theta \cup \Gamma}(\phi) \equiv Cons_{\mathbf{A}}^{\Theta}(\phi_{\Theta}) \wedge Cons_{\mathbf{A}}^{\Gamma}(\phi_{\Gamma}).$$

The theorem is intuitively saying that we can decompose a consequence with respect to a database $\Theta \cup \Gamma$ into two simpler consequences, one with respect to database $\Theta$ and the other with respect to database $\Gamma$, as long the atoms shared by $\Theta$ and $\Gamma$ appear in the observation $\phi$.

Now, what if the observation $\phi$ does not contain all atoms that are shared between $\Theta$ and $\Gamma$? We can still decompose the computation of a consequence $Cons_{\mathbf{A}}^{\Theta \cup \Gamma}(\phi)$ in such a case, but at the expense of performing a case analysis on the shared atoms between $\Theta$ and $\Gamma$. For this we need the following key theorem:

**Theorem 5 (Case-Analysis)** *Let $(\mathbf{P}, \mathbf{A}, \Delta)$ be a system description, $\phi$ be a system observation, and let $\mathbf{S}$ be a subset of $\mathbf{P}$. Then*

$$Cons_{\mathbf{A}}^{\Delta}(\phi) \equiv \bigvee_{\alpha} Cons_{\mathbf{A}}^{\Delta}(\phi \wedge \alpha),$$

*where $\alpha$ ranges over all instantiations of atoms $\mathbf{S}$ that are consistent with $\phi$.*

Case-Analysis is typically used to set the stage for Decomposition. That is, if the observation $\phi$ does not contain all atoms that are shared between $\Theta$ and $\Gamma$, we would perform a case analysis on shared atoms not appearing in $\phi$ and then apply Decomposition to each of the resulting cases. The application of Case-Analysis followed by Decomposition is summarized below:

**Corollary 1 (Intersection)** *Let $(\mathbf{P}, \mathbf{A}, \Theta \cup \Gamma)$ be a system description, $\phi$ be a system observation, and let $\mathbf{S}$ contain all atoms shared by $\Theta$ and $\Gamma$. If $\mathbf{S}$ contains no assumables, then*

$$Cons_{\mathbf{A}}^{\Theta \cup \Gamma}(\phi) \equiv \bigvee_{\alpha} Cons_{\mathbf{A}}^{\Theta}(\alpha \wedge \phi_{\Theta}) \wedge Cons_{\mathbf{A}}^{\Gamma}(\alpha \wedge \phi_{\Gamma}),$$

*where $\alpha$ ranges over all instantiations of $\mathbf{S}$ that are consistent with $\phi$.*

Using Decomposition and Case-Analysis one can always construct a system consequence from component consequences as follows. We partition the component descriptions in $\Delta$ into two subsets, $\Theta$ and $\Gamma$. We identify the common atoms between $\Theta$ and $\Gamma$ and then apply the Intersection Corollary. This allows us to decompose a consequence $Cons^{\Delta}(.)$ into a number of smaller consequences of the form $Cons^{\Theta}(.)$ and $Cons^{\Gamma}(.)$, where each of $\Theta$ and $\Gamma$ are smaller than $\Delta$. We then apply the same procedure, recursively, on each of $Cons^{\Theta}(.)$ and $Cons^{\Gamma}(.)$ until we reach component consequences which can be computed using the Component-Consequence Theorem.

This procedure will always work. However, it does not guarantee the size of the resulting consequence. Specifically, the procedure does not tell us how to partition a database $\Delta$ into two databases $\Theta$ and $\Gamma$. Depending on this choice, the number of common atoms between the partitioned databases will vary; hence, leading to a better or worse consequence size.[9]

As we shall see next, a system structure can play a key role in making this partitioning choice. A system structure is a directed acyclic graph that explicates the interconnections between system components. When a system description is augmented with a system structure, we refer to the result as a structured system description. Structured system descriptions are defined formally in the following section, but first the following example on applying the Intersection Corollary.

Consider Figure 2 where $\Delta = \Delta_C \cup \Delta_D$, and let us compute the consequence of the system observation $\phi = C \wedge D$, that is, $Cons^{\Delta}(\phi)$. We need to construct this consequence from component consequences of the form $Cons^{\Delta_C}(.)$ and $Cons^{\Delta_D}(.)$. We cannot do this immediately, however, since the observation $\phi$ does not mention atom $A$ which is shared by $\Delta_C$ and $\Delta_D$. Therefore, we can use the Intersection Corollary with $\mathbf{S} = \{A\}$:

$$Cons^{\Delta}(\phi) \equiv [Cons^{\Delta_C}(A \wedge \phi_{\Delta_C}) \wedge Cons^{\Delta_D}(A \wedge \phi_{\Delta_D})] \vee [Cons^{\Delta_C}(\neg A \wedge \phi_{\Delta_C}) \wedge Cons^{\Delta_D}(\neg A \wedge \phi_{\Delta_D})].$$

---

9. Realize that the expansion suggested by Case-Analysis is exponential in the number of atoms on which we do the case analysis. Therefore, it is important to choose partitions that will minimize the common atoms between the partitioned databases.

Substituting for $\phi_{\Delta_C} = C$ and $\phi_{\Delta_D} = D$, we get:

$$Cons^{\Delta}(C \wedge D) \equiv [Cons^{\Delta_C}(A \wedge C) \wedge Cons^{\Delta_D}(A \wedge D)] \vee [Cons^{\Delta_C}(\neg A \wedge C) \wedge Cons^{\Delta_D}(\neg A \wedge D)].$$

The resulting expression contains only component consequences, which we assume are pre-computed. Substituting for the values of these consequences we get:

$$Cons^{\Delta}(C \wedge D) \equiv [\neg okX \wedge true] \vee [true \wedge \neg okY] \equiv \neg okX \vee \neg okY.$$

Note that the Intersection Corollary constructs only conjunctions and disjunctions (no negations). Therefore, if component consequences are in NNF, then any system consequence constructed using the corollary must also be in NNF. Moreover, if we ensure that component descriptions do not share assumables, then we are guaranteed that the resulting NNF is also decomposable.[10] This follows because the Intersection Corollary will apply the conjoin operator only to consequences which correspond to disjoint subsystems, that is, collections of disjoint component descriptions.

### 3.3 Structured System Descriptions

We now turn to the formal definition of a structured system description. We start with the definition of a component description (with respect to some assumables $\mathbf{A}$ and non–assumables $\mathbf{P}$).

**Definition 10 (Component Description)** *A component description is a triple $(\mathbf{I}, O, \Delta_O)$ where $\mathbf{I}$ is a set of non–assumables; $O$ is a non–assumable such that $O \notin \mathbf{I}$; and $\Delta_O$ is a set of propositional sentences satisfying the following conditions:*

  1. *$\Delta_O$ can only mention assumables in $\mathbf{A}$ or non–assumables in $\mathbf{I} \cup \{O\}$.*

  2. *Every instantiation of $\mathbf{I} \cup \mathbf{A}$ is consistent with $\Delta_O$.*

The second condition above prohibits a component description from specifying a direct relationship between the inputs of a component and is typically self imposed.

We are now ready to provide the formal definition of a structured system description. We use $\mathcal{G}_P$ to denote the parents of node $P$ in a directed acyclic graph $\mathcal{G}$.

**Definition 11 (Structured System Description)** *A structured system description (SSD) is a tuple $(\mathbf{P}, \mathbf{A}, \mathcal{G}, \Delta)$, where $\mathbf{P}$ and $\mathbf{A}$ are sets of atomic propositions such that $\mathbf{P} \cap \mathbf{A} = \emptyset$; $\mathcal{G}$ is a directed acyclic graph over nodes $\mathbf{P}$; $\Delta$ is a function that maps each node $P$ in $\mathbf{P}$ into a set of propositional sentences $\Delta_P$ such that $(\mathcal{G}_P, P, \Delta_P)$ is a component description. Here, $\mathbf{P}$ is called the set of non–assumables; $\mathbf{A}$ is called the set of assumables; $\mathcal{G}$ is called the system structure. It is required that no assumables be shared between component descriptions.*

---

10. Assuming that component descriptions are themselves decomposable.

We shall overload the meaning of $\Delta$ and use it to denote $\bigcup_{P \in \mathbf{P}} \Delta_P$ (the union of all component descriptions).

The last requirement in Definition 11 ensures the decomposability of generated consequences and does not limit the expressive power of structured system descriptions. Specifically, one can always ensure that component descriptions do not share assumables, but at the expense of adding auxiliary nodes to the system structure. See Appendix C for details.

Each structured system description $(\mathbf{P}, \mathbf{A}, \mathcal{G}, \Delta)$ induces a system description $(\mathbf{P}, \mathbf{A}, \Delta)$ according to Definition 2:

**Theorem 6** *If $(\mathbf{P}, \mathbf{A}, \mathcal{G}, \Delta)$ is a structured system description, then $(\mathbf{P}, \mathbf{A}, \Delta)$ is a system description.*

That is, if $(\mathbf{P}, \mathbf{A}, \mathcal{G}, \Delta)$ is a structured system description, then every $\mathbf{A}$–instantiation is consistent with database $\Delta$. This also means that the database of a structured system description is guaranteed to be consistent by construction. Note that this global consistency is guaranteed given only local conditions on component descriptions.

We close this section by noting that a structured system description as defined here is the diagnosis special-case of a *symbolic causal network* which we introduced elsewhere (Darwiche & Pearl, 1994).

## 4. Structure-Based Computation of Consequences

A main message of the previous section is that composing a system consequence from component consequences is straightforward, as long as we are not concerned about the size of the resulting consequence. Specifically, given that component consequences are precomputed, one can compute a system consequence by successive applications of the Intersection Corollary. In practice, however, the size of the resulting consequence is a key concern because it affects the time needed to generate the consequence, the space needed to store it, and the time needed to extract from it the minimal diagnoses.

Therefore, we shall present in this section a method for composing a system consequence from component consequences while trying to minimize the size of the resulting consequence. This method rests on partitioning component descriptions using a *jointree*: a tree of hypernodes that results from graphically transforming the system structure. In particular,

- we present in Section 4.1 jointrees and show how they can be used to partition component descriptions;

- we then present in Section 4.2 an algorithm for computing component consequences which are the building blocks of system consequences; and

- finally, we present in Section 4.3 an algorithm for computing system consequences using component consequences and a jointree.

### 4.1 Partitioning Component Descriptions using Jointrees

A jointree $\mathcal{T}$ is constructed for a given directed acyclic graph $\mathcal{G}$. The nodes of a jointree are called *clusters* or *cliques* and they represent sets of nodes in the graph $\mathcal{G}$. Figure 5 contains a jointree for the system structure (directed acyclic graph) in Figure 3. The *sepset* of any adjacent cliques $\mathcal{C}_i$ and $\mathcal{C}_j$ in a jointree is defined as their intersection $\mathcal{C}_i \cap \mathcal{C}_j$ and is denoted by $\mathcal{S}_{ij}$.

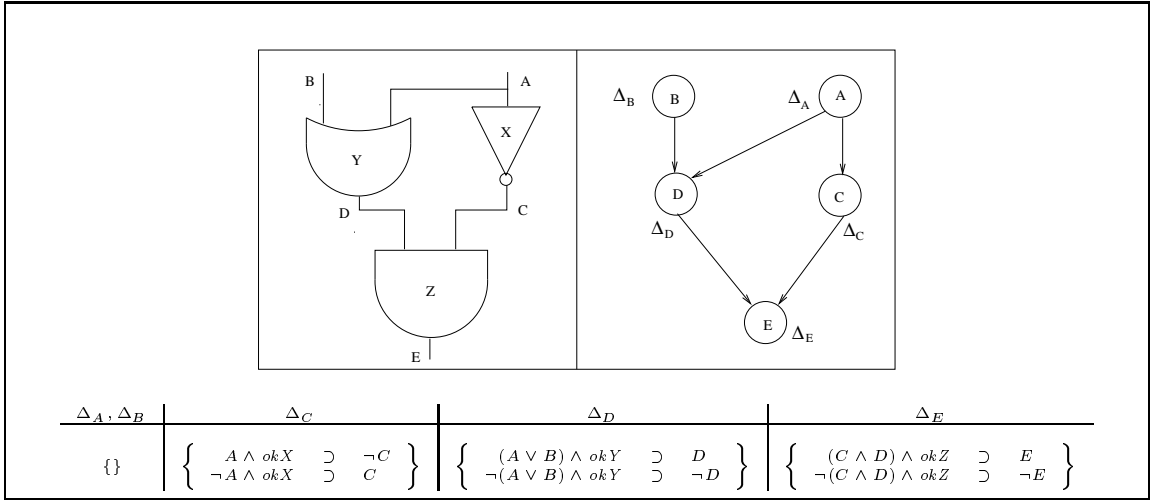

Figure 3: A structured system description of a digital circuit.

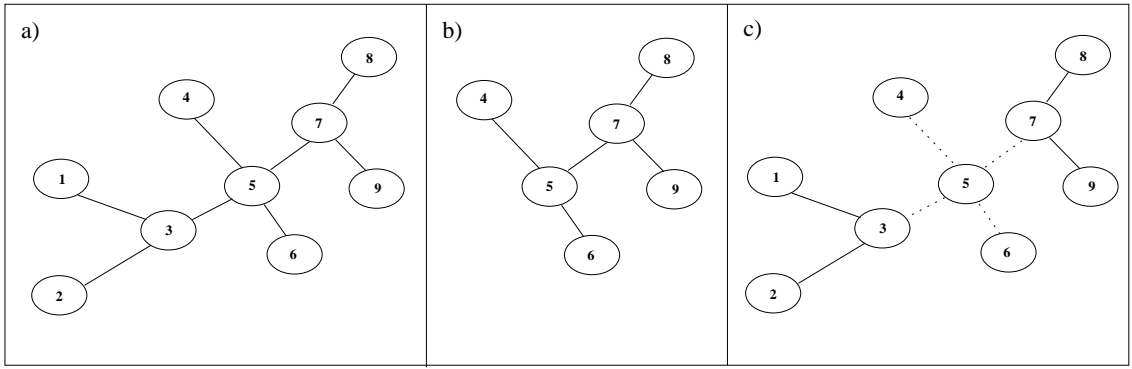

Figure 4: (a) a tree $\mathcal{T}$; (b) a subtree $\mathcal{T}_{53}$; (c) the partitioning of $\mathcal{T}$ around clique $\mathcal{C}_5$.

There are two conditions that must be satisfied by a jointree:

1. The ports of each component must belong to some clique in the jointree.

2. If a node belongs to two cliques, it must also belong to every clique on the path between them. This is called the *jointree property*.

We partition component descriptions using a jointree by assigning each component to a clique that contains the ports of that component. We will use COMPONENTS_OF($\mathcal{C}$) to denote the components assigned to clique $\mathcal{C}$ and refer to the function COMPONENTS_OF as a *component assignment* for $(\mathcal{G}, \mathcal{T})$. Note that a component assignment is not unique. The literature on probabilistic reasoning contains heuristics for choosing a component assignment, but they are outside the scope of this paper (Huang & Darwiche, 1996; Jensen, Lauritzen, & Olesen, 1990).

To see how jointrees are used to generate consequences, we need the following notation. Let $\mathcal{T}$ be a jointree and let $\mathcal{T}_{ij}$ denote the subtree that continues to include clique $\mathcal{C}_i$ after deleting the arc $\mathcal{C}_i$–$\mathcal{C}_j$ from $\mathcal{T}$. For example, Figure 4(a) depicts a jointree $\mathcal{T}$ and Figure 4(b) depicts the subtree $\mathcal{T}_{53}$.

Now, each clique $\mathcal{C}_i$ in a jointree partitions the jointree into clique $\mathcal{C}_i$ and a number of subtrees $\mathcal{T}_{ji}$ where $\mathcal{C}_j$ is a neighbor of $\mathcal{C}_i$. Moreover, by the jointree property, clique $\mathcal{C}_i$ is guaranteed to contain all atoms that are shared by any two subtrees $\mathcal{T}_{ji}$. Therefore, each clique $\mathcal{C}_i$ can be viewed as partitioning components into those assigned to $\mathcal{C}_i$ and those assigned to subtrees $\mathcal{T}_{ji}$. For example, one can partition the jointree in Figure 4(a) into clique $\mathcal{C}_5$ and the subtrees $\mathcal{T}_{35}$, $\mathcal{T}_{45}$, $\mathcal{T}_{65}$ and $\mathcal{T}_{75}$ as shown in Figure 4(c). Moreover, any atom that is shared by two subtrees $\mathcal{T}_{x5}$ and $\mathcal{T}_{y5}$ must belong to clique $\mathcal{C}_5$. Therefore, the atoms of this clique make a very good candidate for being the set $\mathbf{S}$ in the Intersection Corollary; that is, the atoms on which to perform case analysis. If we apply the Intersection Corollary with this choice of $\mathbf{S}$, we will decompose a system consequence (one with respect to the components of $\mathcal{T}$) into a number of consequences some of which are with respect to the components assigned to clique $\mathcal{C}_5$, the others are with respect to the components assigned to subtrees $\mathcal{T}_{35}$, $\mathcal{T}_{45}$, $\mathcal{T}_{65}$ and $\mathcal{T}_{75}$. Each one of the resulting consequences is with respect to a smaller system, and the decomposition process can continue recursively until we reach boundary conditions, where each consequence is with respect to the components assigned to a clique. Clique consequences can be computed easily from the consequences of components assigned to them. More details on this will be given in Section 4.3 where we provide the pseudocode for an algorithm based on this decomposition process.

We shall close this section by explaining why the partition induced by a jointree is better than an arbitrary one. The answer is simple: the optimization criterion for constructing jointrees attempts to minimize the size of cliques. Therefore, the optimization criterion for constructing jointrees attempts to minimize the atoms on which to perform case analysis when applying the Intersection Corollary, therefore, attempting to minimize the size of the resulting consequence. For a detailed, self–contained discussion on the construction of jointrees, the reader is referred to (Huang & Darwiche, 1996).

## 4.2 Computing Component Consequences

In addition to constructing a jointree and a component assignment, we must compute component consequences before we can compute system consequences. In particular, for a given component description $(\mathbf{I}, O, \Delta_O)$, we need to compute $Cons_{\mathbf{A}}^{\Delta_O}(\gamma)$ for every instantiation $\gamma$ of the atoms $\mathbf{I} \cup \{O\}$. That is, for every possible state $\gamma$ of the component ports, $\mathbf{I} \cup \{O\}$, we must compute and store the strongest conclusion, $Cons_{\mathbf{A}}^{\Delta_O}(\gamma)$, we can draw about the health of that component.

This is accomplished by the algorithm in Figure 6 which is a direct implementation of the Component-Consequence Theorem in Section 3.1. The algorithm assumes that each clause $\alpha$ in the component description $\Delta_O$ is decomposed into two parts, one containing

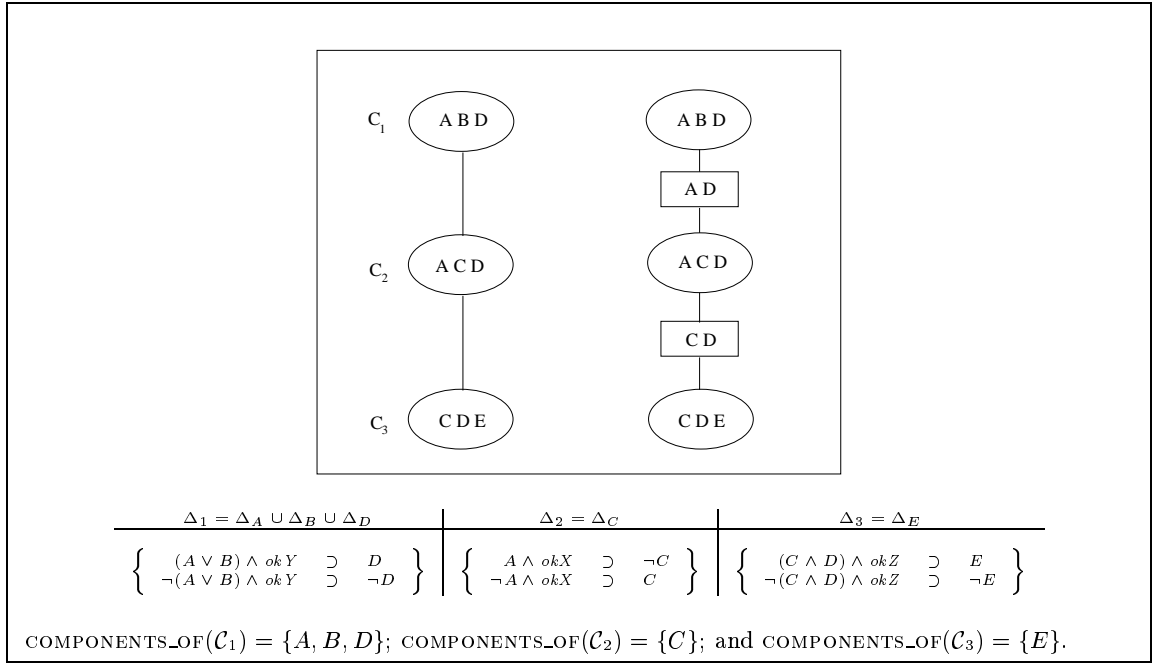

$\Delta_1 = \Delta_A \cup \Delta_B \cup \Delta_D$

$$\left\{ \begin{array}{rcl} (A \vee B) \wedge okY & \supset & D \\ \neg(A \vee B) \wedge okY & \supset & \neg D \end{array} \right\}$$

$\Delta_2 = \Delta_C$

$$\left\{ \begin{array}{rcl} A \wedge okX & \supset & \neg C \\ \neg A \wedge okX & \supset & C \end{array} \right\}$$

$\Delta_3 = \Delta_E$

$$\left\{ \begin{array}{rcl} (C \wedge D) \wedge okZ & \supset & E \\ \neg(C \wedge D) \wedge okZ & \supset & \neg E \end{array} \right\}$$

COMPONENTS_OF($\mathcal{C}_1$) = $\{A, B, D\}$; COMPONENTS_OF($\mathcal{C}_2$) = $\{C\}$; and COMPONENTS_OF($\mathcal{C}_3$) = $\{E\}$.

Figure 5: On the left, a jointree for the structured system description of Figure 3. On the right, the same jointree with sepsets shown on each arc. The table shows the component descriptions, $\Delta_i$, that are assigned to clique $\mathcal{C}_i$.

only assumables, $\alpha_{\mathbf{A}}$, and another containing only non–assumables, $\alpha_{\mathbf{P}}$. It returns an array whose indices correspond to instantiations of the atoms $\mathbf{I} \cup \{O\}$ and whose entries are component consequences.

Please note that we use integers to represent instantiations, a technique that we detail in Appendix B. This appendix provides two key functions, one for computing a unique index for each instantiation and another for generating all instantiations of a given set of atoms. These functions are used in the pseudocode of Figure 6.

The soundness of the algorithm in Figure 6 is given below:

**Theorem 7** *The array* CONSEQUENCES$_O$ *computed in Figure 6 satisfies the following property:*

$$\text{CONSEQUENCES}_O[l] \equiv Cons_{\mathbf{A}}^{\Delta_O}(\gamma)$$

*for every instantiation $\gamma$ of $\mathbf{I} \cup \{O\}$ and its corresponding index $l$.*

Figure 7 provides a detailed example showing how the algorithm of Figure 6 is used to compute the consequences of a 2-input or-gate. The number of computed consequences in this example is eight, which is $2^{2+1}$, since the component has two inputs and one output. Each component consequence, $Cons_{\mathbf{A}}^{\Delta_D}(\gamma)$, in this example is equivalent to either *true* or $\neg ok\,Y$ depending on whether the instantiation $\gamma$ of atoms $\{A, B, D\}$ is consistent with the expected behavior of the or-gate.

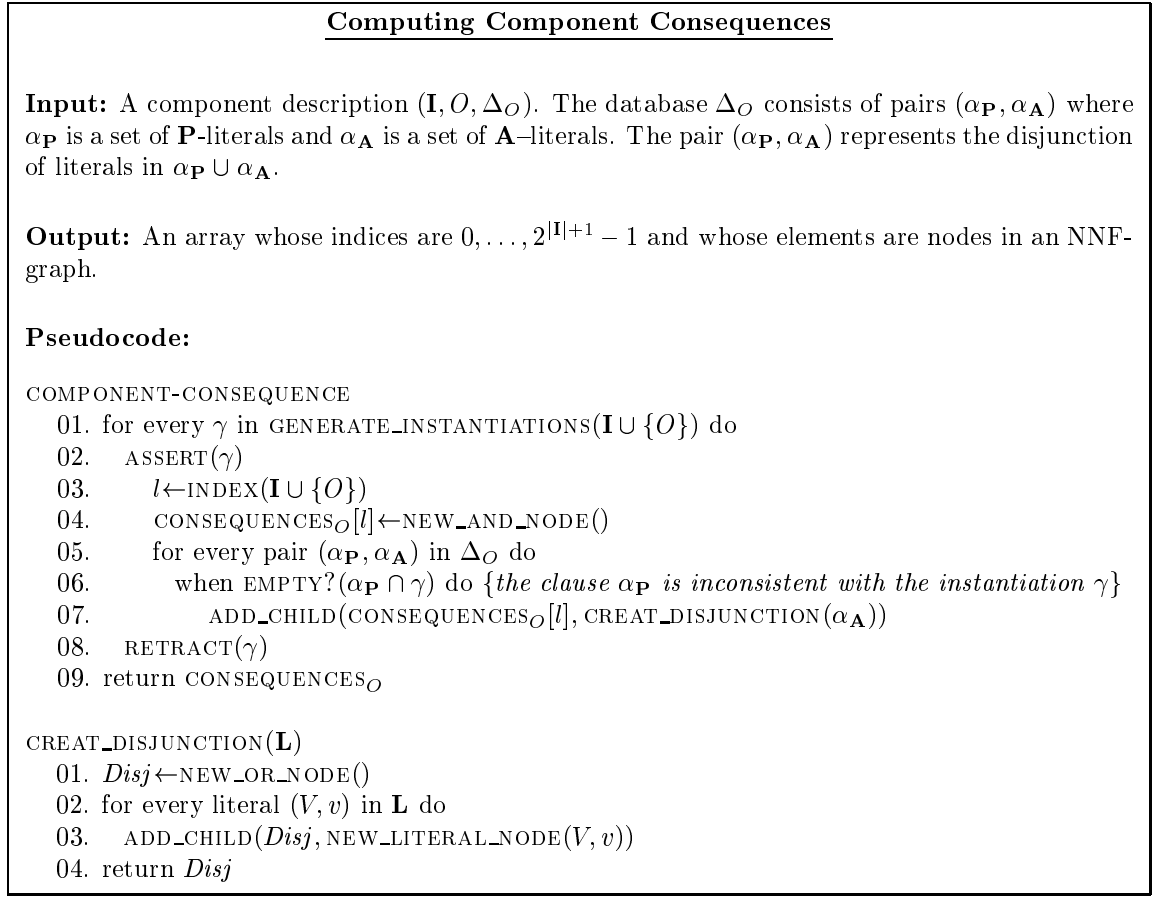

Figure 6: An algorithm for computing component consequences.

The following theorem shows that computing component consequences is exponential in the number of component ports, but is linear in the number of clauses in its description.

**Theorem 8** *The time and space complexity of* COMPONENT_CONSEQUENCE *is* $O(sn2^n)$ *where $s$ is the number of pairs in database $\Delta_O$ and $n$ is the number of atoms in $\mathbf{I} \cup \{O\}$.*

Therefore, as long as the number of inputs to a component is small enough, computing component consequences can be considered to take constant time. Given this complexity result, one should attempt to minimize the number of inputs per component. Bear in mind that components are conceptual constructs and need not strictly correspond to physical components in the system. For example, one may opt to view an $n$-input and-gate as a number of cascaded 2-input and-gates. This will reduce the complexity of computing component consequences from $O(sn2^n)$ to $O(s)$; a significant improvement.[11]

## 4.3 Computing System Consequences

Given that we have

- constructed a jointree;

- chosen a component assignment;

- computed component consequences;

we are now ready to compute system consequences. All we need is to apply the Intersection Corollary successively until we decompose the system consequence into a number of component consequences which can be simply looked-up from the arrays that are computed by the algorithm of Figure 6.

To apply the Intersection Corollary, we need a pivot clique which is given as input to the algorithm in Figure 8. The algorithm has two main functions, one for computing a consequence with respect to the components assigned to a clique and another for computing a consequence with respect to the components assigned to a subtree. Let us consider the function for computing clique consequences first:

**Lemma 1** When CLIQUE_CONSEQUENCE($\mathcal{C}_i$) is called, atoms $\mathcal{C}_i$ are guaranteed to be instantiated to some $\alpha$ and the call returns a sentence equivalent to $Cons_{\mathbf{A}}^{\Delta_i}(\phi_i \wedge \alpha)$ where

- $\Delta_i$ is the union of all component descriptions assigned to clique $\mathcal{C}_i$; and

- $\phi_i$ is the projection of system observation $\phi$ on the atoms in clique $\mathcal{C}_i$.

This function is simple; it directly applies the Decomposition theorem to decompose a clique consequence into the conjunction of component consequences (which are provided as input to the algorithm).

The second function of the algorithm computes a consequence with respect to the components assigned to a subtree $\mathcal{T}_{ij}$:

**Lemma 2** When SUBTREE_CONSEQUENCE($\mathcal{C}_i, \mathcal{C}_j$) is called, atoms $\mathcal{S}_{ij}$ are guaranteed to be instantiated to some $\beta$ and the call returns a sentence equivalent to $Cons_{\mathbf{A}}^{\Delta_{ij}}(\phi_{ij} \wedge \beta)$ where

- $\Delta_{ij}$ is the union of all component descriptions assigned to cliques in subtree $\mathcal{T}_{ij}$; and

- $\phi_{ij}$ is the projection of system observation $\phi$ on the atoms in subtree $\mathcal{T}_{ij}$.

This function is more involved since it calls CLIQUE_CONSEQUENCE and itself recursively.

Note that the results returned by this function are cached since the function may be called more than once with the same arguments and the same instantiation of sepset $\mathcal{S}_{ij}$.[12]

---

11. In such a case, however, the technique that we discuss in Appendix C must be used to ensure that the descriptions of these 2-input components do not share assumables.

12. This is because the same instantiation of a sepset $\mathcal{S}_{ij}$ may appear as part of many instantiations of clique $\mathcal{C}_i$.

---

**An Example of Algorithm** COMPONENT_CONSEQUENCE

Consider the following component description, $(\{A, B\}, D, \Delta_D)$, of a 2-input or-gate where

$$\Delta_D = \left\{ \begin{array}{c} \neg A \wedge \neg B \wedge okY \supset \neg D \\ A \wedge okY \supset D \\ B \wedge okY \supset D \end{array} \right\}.$$

Algorithm COMPONENT_CONSEQUENCE requires the following representation of $\Delta_D$:

$$\left\{ \begin{array}{c} (\{(A, 1), (B, 1), (D, 0)\}, \{(okY, 0)\}) \\ (\{(A, 0), (D, 1)\}, \{(okY, 0)\}) \\ (\{(B, 0), (D, 1)\}, \{(okY, 0)\}) \end{array} \right\},$$

where each clause is represented as a pair of clauses, one containing only non-assumables and the other containing only assumables. Calling COMPONENT_CONSEQUENCE on this component description returns the following array (first two columns):

| index | CONSEQUENCES$_D$ | $\gamma$ | $Cons_{\mathbf{A}}^{\Delta_D}(\gamma)$ |
|-------|------------------|----------|----------------------------------------|
| 0 | 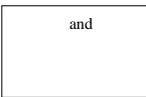 | $\neg A \wedge \neg B \wedge \neg D$ | $true$ |
| 1 | 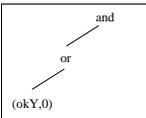 | $A \wedge \neg B \wedge \neg D$ | $\neg okY$ |
| 2 | 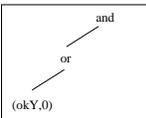 | $\neg A \wedge B \wedge \neg D$ | $\neg okY$ |
| 3 | 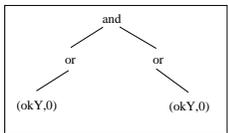 | $A \wedge B \wedge \neg D$ | $\neg okY$ |
| 4 | 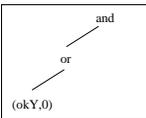 | $\neg A \wedge \neg B \wedge D$ | $\neg okY$ |
| 5 | 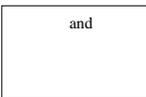 | $A \wedge \neg B \wedge D$ | $true$ |
| 6 | 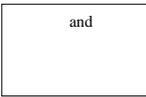 | $\neg A \wedge B \wedge D$ | $true$ |
| 7 | 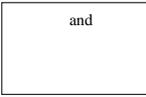 | $A \wedge B \wedge D$ | $true$ |

Figure 7: Computing consequenes of an or-gate with inputs $\{A, B\}$ and output $D$.

---

**Computing System Consequences**

**Input:**

- $(\mathbf{P}, \mathbf{A}, \mathcal{G}, \Delta)$: a structured system description;
- $\phi$: an instantiation of some atoms $\mathbf{E} \subseteq \mathbf{P}$;
- $\mathcal{T}$: a join tree for system structure $\mathcal{G}$;
- COMPONENTS_OF: a component assignment for $(\mathcal{G}, \mathcal{T})$;
- CONSEQUENCES$_O$: a component consequence array for each component $O$ — computed using the algorithm in Figure 6;
- $\mathcal{C}_z$: an arbitrary cluster in the tree $\mathcal{T}$ (pivot cluster).

**Output:** An NNF-graph over assumables $\mathbf{A}$.

**Data Structures:** For each neighboring clusters $(\mathcal{C}_x, \mathcal{C}_y)$ in the join tree $\mathcal{T}$:

- a sepset $\mathcal{S}_{xy} = \mathcal{C}_x \cap \mathcal{C}_y$; and

- a hash table HASH_TABLE$_{xy}$ which contains NNF-graph nodes.

The keys of HASH_TABLE$_{xy}$ are $0, \ldots, 2^{|\mathcal{S}_{xy}|} - 1$. HASH_TABLE$_{xy}[l]$ is the element having $l$ as a key.

**Pseudocode:**

SYSTEM_CONSEQUENCE
    01. add cluster $\mathcal{C}_0 = \emptyset$ as a neighbor of $\mathcal{C}_z$
    02. ASSERT($\phi$)
    03. SUBTREE_CONSEQUENCE($\mathcal{C}_z, \mathcal{C}_0$)

SUBTREE_CONSEQUENCE($\mathcal{C}_i, \mathcal{C}_j$)
    00. $l \leftarrow$ INDEX($\mathcal{S}_{ij}$)
    01. $\mu \leftarrow$ HASH_TABLE$_{ij}[l]$; if $\mu \neq$ NIL then return $\mu$
    02. $Disj \leftarrow$ NEW_OR_NODE()
    03. for every $\alpha$ in GENERATE_INSTANTIATIONS($\mathcal{C}_i$) do
    04.    ASSERT($\alpha$)
    05.    $Conj \leftarrow$ NEW_AND_NODE()
    06.    ADD_CHILD($Conj$, CLIQUE_CONSEQUENCE($\mathcal{C}_i$))
    07.    for every neighboring cluster $\mathcal{C}_k$ of $\mathcal{C}_i$ where $k \neq j$ do
    08.      ADD_CHILD($Conj$, SUBTREE_CONSEQUENCE($\mathcal{C}_k, \mathcal{C}_i$))
    09.    ADD_CHILD($Disj$, $Conj$)
    10.    RETRACT($\alpha$)
    11. HASH_TABLE$_{ij}[l] \leftarrow Disj$; return $Disj$

CLUSTER_CONSEQUENCE($\mathcal{C}_i$)
    01. $Conj \leftarrow$ NEW_AND_NODE()
    02. for every component $O$ (with inputs $\mathbf{I}$) in COMPONENTS_OF($\mathcal{C}_i$) do
    03.    ADD_CHILD($Conj$, CONSEQUENCES$_O$[INDEX($\mathbf{I} \cup \{O\}$)])
    04. return $Conj$

Figure 8: An algorithm for computing a system consequence given a system observation.

In such a case, the result is looked up from the cache instead of being recomputed. This caching of results explains why the generated consequence has the form of a directed acyclic graph instead of a tree.

Given the function SUBTREE_CONSEQUENCE, one can compute the system consequence by adding a dummy clique $\mathcal{C}_0 = \emptyset$ as a neighbor to the pivot clique $\mathcal{C}_z$ to generate an extended jointree $\mathcal{T}'$. The subtree $\mathcal{T}'_{z_0}$ of this extended tree $\mathcal{T}'$ will be nothing but the original jointree $\mathcal{T}$. The call to SUBTREE_CONSEQUENCE on line 03 of SYSTEM_CONSEQUENCE will then return the system consequence. This is stated more explicitly by the following theorem:

**Theorem 9** *The function* SYSTEM_CONSEQUENCE *in Figure 8 will always terminate returning a sentence which is equivalent to* $Cons_{\mathbf{A}}^{\Delta}(\phi)$.

Figure 9 contains a detailed example showing how the algorithm in Figure 8 is used to generate a system consequence for a given observation and structured system description. Note that the computed consequence in this example is equivalent to $\neg okX \vee \neg okZ$.

The consequences computed by the algorithm in Figure 8 are guaranteed to be in decomposable negation normal form as long as component consequences are in that form.

**Theorem 10** *In Figure 8, if each component consequence* CONSEQUENCES$_O$[.] *is in decomposable NNF, the sentence returned by the function* SYSTEM_CONSEQUENCE *will be in decomposable NNF.*

Note that the consequences computed by COMPONENT_CONSEQUENCE in Figure 6 are not guaranteed to be in decomposable NNF. However, they are guaranteed to be in CNF. We can simply transform them into DNF and, hence, into decomposable NNF.

## 5. The Complexity of Computing Consequences

We now turn to the discussion of computational complexity:

**Theorem 11** *Consider the algorithm* SYSTEM_CONSEQUENCE *in Figure 8. The time and space complexity of this algorithm is*

$$O(\sum_{\mathcal{C} \in \mathcal{T}} \mid \mathcal{C} \mid 2^{|\mathcal{C} \setminus \mathbf{E}|}),$$

*where* $\mathcal{C}$ *denotes a clique in the jointree* $\mathcal{T}$.

The complexity is then linear in the number of cliques, but exponential in their sizes.

Consider all jointrees of a given graph $\mathcal{G}$, and select the tree $\mathcal{T}$ with the smallest maximal clique. The size of the maximal clique in such a tree minus one is known as the width $w^*$ of graph $\mathcal{G}$ (Dechter, 1992; Dechter & Pearl, 1989). Theorem 11 is then saying that the computational complexity of a consequence is exponential in the width $w^*$ of the given system structure $\mathcal{G}$. But this complexity result is for the worst case — the best and average cases are different.

## An Example of Algorithm SYSTEM_CONSEQUENCE

Consider the jointree and component assignment given in Figure 5. Consider also the following component consequence arrays:

| $\gamma$ | | CONSEQUENCES$_C$ |
|---|---|---|
| $\neg A \wedge \neg C$ | 0 | $(okX, 0)$ |
| $A \wedge \neg C$ | 1 | $true$ |
| $\neg A \wedge C$ | 2 | $true$ |
| $A \wedge C$ | 3 | $(okX, 0)$ |

| $\gamma$ | | CONSEQUENCES$_D$ |
|---|---|---|
| $\neg A \wedge \neg B \wedge \neg D$ | 0 | $true$ |
| $A \wedge \neg B \wedge \neg D$ | 1 | $(okY, 0)$ |
| $\neg A \wedge B \wedge \neg D$ | 2 | $(okY, 0)$ |
| $A \wedge B \wedge \neg D$ | 3 | $(okY, 0)$ |
| $\neg A \wedge \neg B \wedge D$ | 4 | $(okY, 0)$ |
| $A \wedge \neg B \wedge D$ | 5 | $true$ |
| $\neg A \wedge B \wedge D$ | 6 | $true$ |
| $A \wedge B \wedge D$ | 7 | $true$ |

| $\gamma$ | | CONSEQUENCES$_E$ |
|---|---|---|
| $\neg C \wedge \neg D \wedge \neg E$ | 0 | $true$ |
| $C \wedge \neg D \wedge \neg E$ | 1 | $true$ |
| $\neg C \wedge D \wedge \neg E$ | 2 | $true$ |
| $C \wedge D \wedge \neg E$ | 3 | $(okZ, 0)$ |
| $\neg C \wedge \neg D \wedge E$ | 4 | $(okZ, 0)$ |
| $C \wedge \neg D \wedge E$ | 5 | $(okZ, 0)$ |
| $\neg C \wedge D \wedge E$ | 6 | $(okZ, 0)$ |
| $C \wedge D \wedge E$ | 7 | $true$ |

Calling SYSTEM_CONSEQUENCE with the observation $\phi$ as $\{(A, 1), (E, 1)\}$ and the pivot cluster as $\mathcal{C}_1$, we get the following consequence (in decomposable negation normal form):

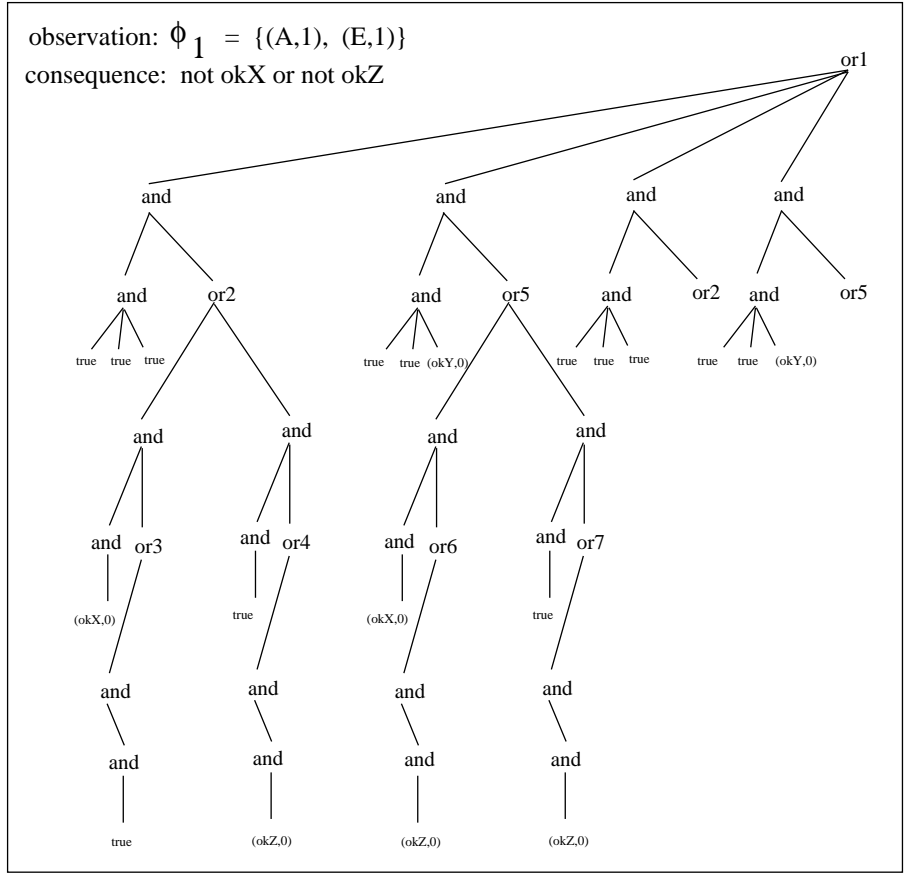

The resulting consequence can be simplified significantly and easily. One way to realize this simplification is to use an optimized version of ADD-CHILD in Appendix A which performs the necessary simplifications as it adds a child to a node.

Figure 9: An example showing the application of algorithm SYSTEM_CONSEQUENCE. We have numbered or-nodes in the consequence to facilitate the depiction of structure sharing without having to draw cross arcs.

## 5.1 The Effect of Observations

According to Theorem 11, the time and space complexity are exponential not in the size of cliques, but in the size of cliques after we have removed from them atoms **E** that appear in the observation $\phi$. At one extreme, the observation $\phi$ is *true* (**E** = $\emptyset$) in which case the algorithm is exponential in the size of the maximal clique. At another extreme, the observation $\phi$ is an instantiation of all non–assumables (**E** = **P**). In such a case, the algorithm does not depend on the size of cliques and is only linear in the number of such cliques. This can be seen by observing line 02 of algorithm SYSTEM_CONSEQUENCE where all atoms in the system structure are instantiated. This means that the result of any call to GENERATE_INSTANTIATIONS will return a singleton. Hence, every constructed or-node on line 02 of SUBTREE_CONSEQUENCE will have exactly one child. Moreover, the number of calls to SUBTREE_CONSEQUENCE will be exactly the number of edges in the jointree and the number of calls to CLIQUE_CONSEQUENCE will be exactly the number of cliques in the jointree.

The effect of observations on the computational complexity of consequences is illustrated in Figure 10 which depicts four consequences with respect to the system in Figure 3. The consequences correspond to the following observations:

- $\phi_1 = A \wedge E$;
- $\phi_2 = A \wedge B \wedge E$;
- $\phi_3 = A \wedge B \wedge C \wedge E$; and
- $\phi_4 = A \wedge B \wedge C \wedge \neg D \wedge E$.

Clearly, observation $\phi_i$ is stronger than observation $\phi_{i-1}$. Interestingly enough, the consequence of $\phi_i$ is a subset of the consequence of $\phi_{i-1}$. Moreover, the subset is obtained by eliminating some children of or-nodes, which corresponds to considering fewer cases when applying Case-Analysis.[13]

Because of this dependence on observations, the presented algorithm for generating consequences will do very well in applications where the system is strongly observed. We have experienced this ourselves when we applied this framework to discrete event systems which are typically strongly observed (Darwiche & Provan, 1996, 1997). Even when the system is not strongly observed, Theorem 11 is useful in deciding which observations would be most rewarding computationally.

A more dramatic effect of the system observation on complexity is obtained by cutting outgoing arcs of observed nodes from the system structure. This technique may actually reduce the $w^*$ of the given system structure and is illustrated in Appendix E, together with some experimental results that show its effectiveness.

## 5.2 Tree System Structures

An important special case of the presented algorithm is when the system structure $\mathcal{G}$ that we start with is already a tree — that is, only one undirected path exists between any two nodes in $\mathcal{G}$ (see Figure 11). In this case, one can always construct a jointree in which (a) each clique contains only a node and its parents in $\mathcal{G}$; and (b) each sepset contains only one node. In particular, for each node $N$ in the system structure $\mathcal{G}$, we construct a clique which contains the family of $N$ (that is, $\{N\} \cup \mathcal{G}_N$). We then connect the clique corresponding to $N$ with the cliques corresponding its neighbors, eliminating any cliques that are contained in their neighbors. The result is guaranteed to be a jointree. Figure 11 contains a jointree of a tree system structure. More details on this construction can be found in (Shachter, Andersen, & Szolovits, 1994).

---

13. This observation is the basis of an approach for compiling devices into parameterized NNFs, which is described in (Darwiche, 1998a).

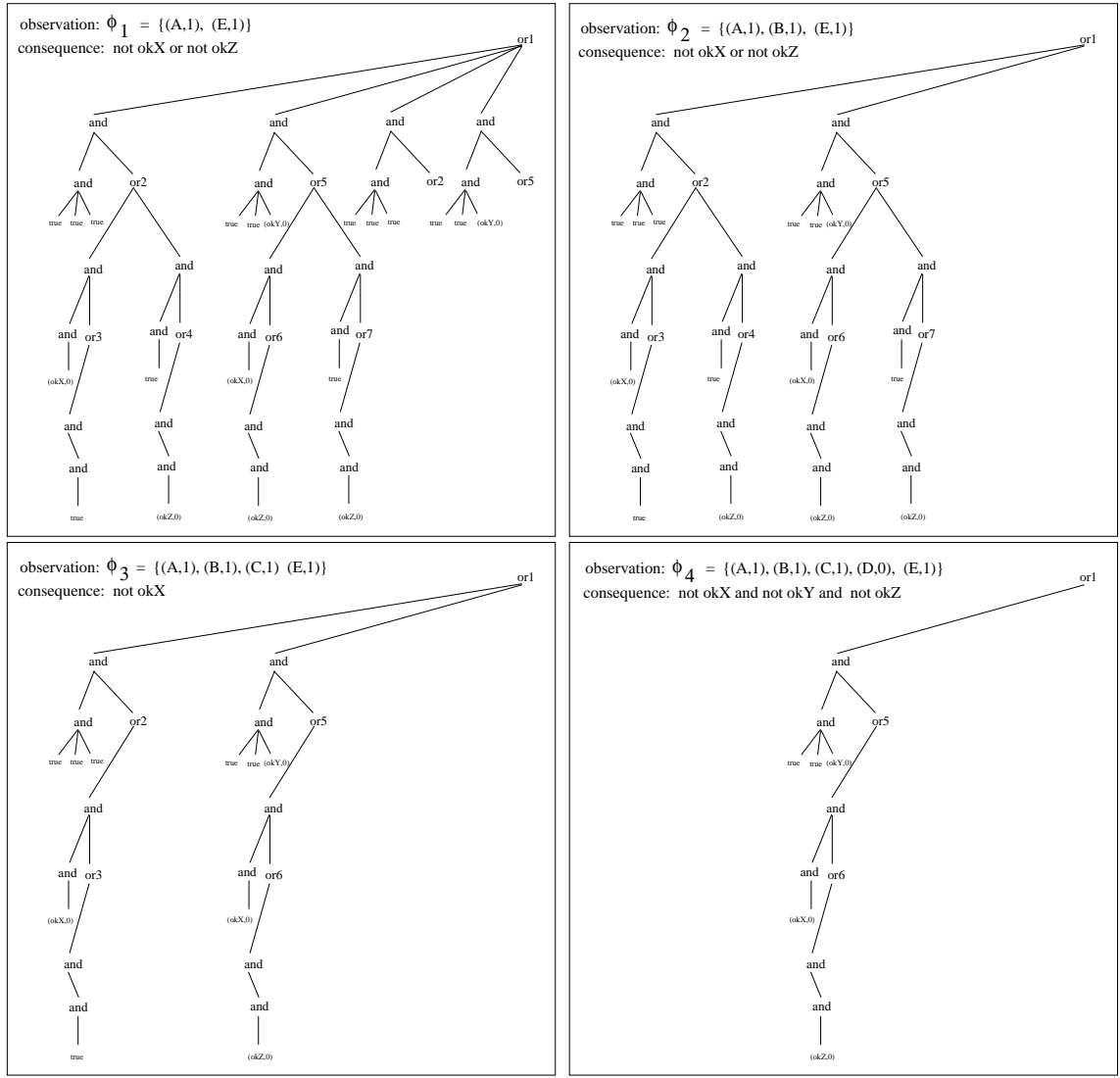

Figure 10: Four consequences corresponding to four different observations.

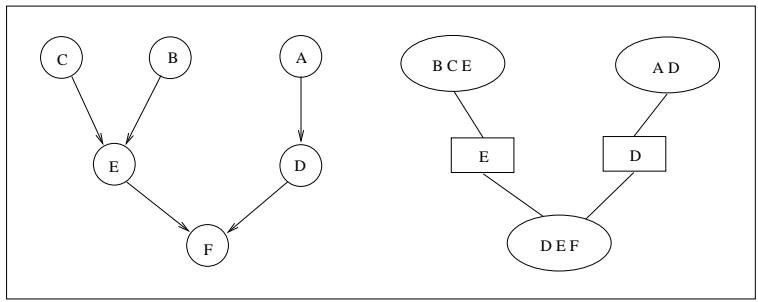

Figure 11: On the left, a system structure with no cycles. On the right, a corresponding jointree.

Therefore, when the system structure is a tree, the time and space complexity of computing a consequence are exponential only in the size of a family. Because of this result, one should attempt to construct tree system structures whenever possible. Moreover, special care must be exercised to avoid large families in such structures. For example, one should model $n$-input gates as a cascaded set of 2-input gates whenever possible. If a tree system structure is not possible, one should attempt to engineer the structure so as to minimize the size of cliques in its corresponding jointrees.

Although tree system structures are computationally easy, non-tree structures are not necessarily hard (El Fattah & Dechter, 1995). In fact, it is possible to obtain a linear complexity even if the system structure is not a tree. Consider for example $n$-bit adders which are constructed form cascading full adders. These adders have non-tree structures, but it is shown in (El Fattah & Dechter, 1995) that they have jointrees which grow linearly in the number of bits $n$. Therefore, the consequences corresponding to these systems grow linearly in the number of bits $n$.

## 6. Extracting Minimal Diagnoses

In this section, we present a method for extracting the minimal diagnoses characterized by a consequence. In particular, if the consequence is in decomposable NNF, and if the minimality criterion satisfies some general conditions, the algorithm we shall present will extract minimal diagnoses in time linear in the size of the consequence.

We start first by stating the required conditions on a minimality criterion and then introduce the algorithm with some examples. We finally state the algorithm formally, prove its soundness, and its linear computational complexity.

## 6.1 Cost Functions

Intuitively, minimal diagnoses are those diagnoses which are considered most plausible. One may have many instantiations of assumables that are consistent with the system description and observation, but only a few of these may be considered most plausible. A common minimality criterion is the one based on diagnosis cardinality. That is, we simply count the number of $\neg ok$. literals in a diagnosis (cardinality of a diagnosis) and consider the diagnoses with minimal cardinality to be the most plausible.

We adopt the view that each diagnosis has a cost, and minimal diagnoses are those with a minimal cost. Moreover, we assume that the cost of a diagnosis is obtained by adding up the costs of its literals. This is all captured by the following notion of a cost function, which assigns costs to literals and instantiations.

**Definition 12** *A* <u>*cost function*</u> *is a tuple* $(\mathbf{A}, \Sigma, \oplus, \mathcal{F})$ *where*

- $\mathbf{A}$ *is a set of atoms;*

- $\Sigma$ *is a set of costs;*

- $\oplus$ *is a binary operation (addition) on* $\Sigma$ *which satisfies the following properties:*

    1. $\oplus$ *is commutative, associative, and has a zero element 0.*
    2. *For all a and b in* $\Sigma$*, either* $a \oplus c = b$ *or* $b \oplus c = a$ *for some unique c.*
    3. $a \oplus b = 0$ *implies* $a = 0$ *and* $b = 0$*.*

- $\mathcal{F}$ *maps each* $\mathbf{A}$*-literal into a cost in* $\Sigma$*, where each literal or its negation has cost 0.*

The function $\mathcal{F}$ is extended to instantiations as follows:

$$\mathcal{F}(l_1 \wedge \ldots \wedge l_n) = \mathcal{F}(l_1) \oplus \ldots \oplus \mathcal{F}(l_n).$$

The conditions we impose on the cost addition function are justified as follows. The first condition is to be expected of addition operations. The second condition ensures that costs are totally ordered; that is, for any two costs, one can be obtained from the other by adding some cost to it. The third condition ensures that there are no negative costs so that we cannot reduce the cost of an instantiation by adding more literals.

A cost function induces an ordering on costs as follows:

**Definition 13** *The ordering relation* $\leq_{\oplus}$ *induced by a cost function* $(\mathbf{A}, \Sigma, \oplus, \mathcal{F})$ *is defined as follows:* $a \leq_{\oplus} b$ *iff* $a \oplus c = b$ *for some c. If* $a \leq_{\oplus} b$ *and* $a \neq b$*, we write* $a <_{\oplus} b$*.*

That is, if $b$ is obtained by adding some cost to $a$, then $b$ is greater than or equal to $a$. Moreover, this induced ordering is guaranteed to be a total ordering by virtue of the properties we imposed on cost addition:

**Theorem 12** *The relation* $\leq_{\oplus}$ *induced by a cost function* $(\mathbf{A}, \Sigma, \oplus, \mathcal{F})$ *is a total ordering on* $\Sigma$*.*

An example of a cost function satisfying the above conditions is $(\{0, 1, 2, \ldots\}, +, \kappa)$, where the cost of a literal $l$, $\kappa(l)$, is the order-of-magnitude of its probability.[14]

Another example of a cost function is the minimum cardinality criterion, which is widely used in the diagnosis literature. The criterion is specified by $(\{0, 1, 2, \ldots\}, +, Card)$, where $Card(l) = 1$ if the literal $l$ designate a fault and $Card(l) = 0$ otherwise.

Note that the triple $([0, 1], *, Pr)$, where the cost of a literal is its probability, is not a legitimate cost function according to the conditions stated above. The cost addition function $*$ has 1 as its zero element in this case. Therefore, the condition $Pr(l) = 1$ or $Pr(\neg l) = 1$ does not necessarily hold for each literal $l$. Note also that the set-inclusion version of diagnosis minimality is not covered by our definition of cost functions; this notion, however, is problematic when fault models are included (de Kleer et al., 1992).

The reason we require each literal or its negation to have cost 0 is to be able to characterize minimal diagnoses using minimal partial diagnoses, a property which is not possible in general. For example, let $\alpha = \alpha_1 \vee \ldots \vee \alpha_n$ be a disjunction of partial diagnoses and let $\alpha'$ be the result of removing from $\alpha$ all partial diagnoses that do not have minimal cost. The above requirement guarantees that $\alpha$ and $\alpha'$ characterize the same set of minimal diagnoses.[15]

## 6.2 The Extraction Algorithm

The goal of the algorithm that we shall present next is to compute diagnoses that are characterized by a consequence $\tau$ and that have a minimal cost compared to other diagnoses characterized by $\tau$.

**Definition 14 (Minimal Instantiations)** *Let $(\mathbf{A}, \Sigma, \oplus, \mathcal{F})$ be a cost function and let $\tau$ be an $\mathbf{A}$-sentence. A minimal instantiation of $\tau$ with respect to $(\mathbf{A}, \Sigma, \oplus, \mathcal{F})$ is an $\mathbf{A}$-instantiation $\alpha$ such that*

1. *$\alpha \models \tau$; and*

2. *if $\beta$ is an $\mathbf{A}$-instantiation such that $\beta \models \tau$, then $\mathcal{F}(\alpha) \leq_\oplus \mathcal{F}(\beta)$.*

*The minimal instantiations of $\tau$ are denoted by $MinInst(\tau)$.*

Consider the consequence $\tau$ in Figure 9 for an example. This consequence is equivalent to $\neg okX \vee \neg okZ$ and, therefore, characterizes six diagnoses:

---

14. The order-of-magnitude of a probability $p$, written $\kappa(p)$, is an integer such that $p/\epsilon^{\kappa(p)}$ is finite but not infinitesimal for an infinitesimal $\epsilon$ (Darwiche & Goldszmidt, 1994; Goldszmidt & Pearl, 1992; Spohn, 1987). That is, $\epsilon^{\kappa(p)+1} < p \leq \epsilon^{\kappa(p)}$. Note that if $a$ and $b$ are probabilistically independent, then $\kappa(Pr(ab)) = \kappa(Pr(a)) + \kappa(Pr(b))$.

15. Consider the following counterexample with respect to the triple $([0, 1], *, Pr)$ which does not satisfy the above requirement. Suppose that the assumables are $okX$, $okY$, $okZ$ and $okW$ and their probabilities are .9, .9, .7 and .9, respectively. Suppose further that $\alpha = okX \wedge okY \wedge \neg okW \vee \neg okY \wedge okZ$. The first partial diagnosis is less probable than the second and, therefore, pruning gives $\alpha' = \neg okY \wedge okZ$. Note, however, that $\alpha$ characterizes six diagnoses, two of which $okX \wedge okY \wedge okZ \wedge \neg okW$ and $okX \wedge \neg okY \wedge okZ \wedge okW$ are most probable. On the other hand, $\alpha'$ characterizes four diagnoses, only one of which $okX \wedge \neg okY \wedge okZ \wedge okW$ is most probable. Therefore, $\alpha$ and $\alpha'$ do not characterize the same set of most probable diagnoses.

1. $okX \wedge okY \wedge \neg okZ$
2. $okX \wedge \neg okY \wedge \neg okZ$
3. $\neg okX \wedge okY \wedge okZ$
4. $\neg okX \wedge okY \wedge \neg okZ$
5. $\neg okX \wedge \neg okY \wedge okZ$
6. $\neg okX \wedge \neg okY \wedge \neg okZ$.

According to the minimum-cardinality cost function where $\mathbf{A} = \{okX, okY, okZ\}$, the costs of these diagnoses are 1, 2, 1, 2, 2, 3, respectively. Therefore, the consequence characterizes only two minimal diagnoses:

1. $okX \wedge okY \wedge \neg okZ$
3. $\neg okX \wedge okY \wedge okZ$.

Hence,

$$MinInst(\tau) = \{okX \wedge okY \wedge \neg okZ, \neg okX \wedge okY \wedge okZ\}.$$

The purpose of our algorithm is to compute this set of diagnoses given consequence $\tau$.

As it turns out, extracting minimal diagnoses from a decomposable negation normal form is quite simple and can be carried out in two steps. In the first step, the consequence is pruned to result in a sentence that characterizes the minimal diagnoses only. In the second step, all instantiations of the pruned consequence are generated, each of which will then be a minimal diagnosis. The algorithm for extracting minimal diagnoses is given in Figure 12 and is explained in more detail below:

## Phase I: Cost Propagation and Pruning

Each node in the NNF is assigned a cost as follows:

1. The cost of a literal is given by the cost function;

2. The cost of a conjunction is the $\oplus$-sum of costs assigned to its conjuncts;

3. The cost of a disjunction is the minimum of costs assigned to its disjuncts.

It is shown by Lemma 7 in Appendix F that the cost assigned to node $\mu$ is nothing but the cost of the minimal instantiations it characterizes.

In addition to this cost propagation, some pruning takes place in this phase. In particular, if a disjunction $\mu$ has a disjunct $\nu$ with a higher cost than $\mu$, the disjunct $\nu$ is removed from $\mu$.[16] The intuition here is that $\nu$ will be contributing partial diagnoses that are not minimal when compared to the partial diagnoses contributed by its siblings. Moreover, all completions of these partial diagnoses will be more costly than some completion of a partial diagnosis contributed by a sibling.

---

16. This is achieved by deleting the link between $\mu$ and $\nu$ in the NNF-graph.

---

### Computing Minimal Instantiations

**Input:**

- $(\mathbf{A}, \Sigma, \oplus, \mathcal{F})$: a cost function;
- $\tau$: a decomposable NNF over assumables $\mathbf{A}$.
- $Atoms(\mu, \nu)$: atoms that appear in $\mu$ but not in $\nu$ for each or-node $\mu$ and its child $\nu$.

**Output:** A set of $\mathbf{A}$-instantiations.

**Data Structures:** For each node $\mu$ in $\tau$, two fields COST$(\mu)$ and TERMS$(\mu)$ initialized to NIL.

**Pseudocode:**

MINIMAL_INSTANTIATIONS$(\tau)$
  01. PRUNE$(\tau)$
  02. INSTANTIATIONS$(\tau)$

PRUNE$(\tau)$
  01. when COST$(\tau)$ = NIL do
  02.     for each $\mu \in$ CHILDREN$(\tau)$ do PRUNE$(\mu)$
  03.     case LITERAL_NODE?$(\tau)$: COST$(\tau) \leftarrow \mathcal{F}($LITERAL_OF$(\tau))$
  04.         AND_NODE?$(\tau)$    : COST$(\tau) \leftarrow \displaystyle\bigoplus_{\mu \in \text{CHILDREN}(\tau)}$ COST$(\mu)$
  05.         OR_NODE?$(\tau)$    : COST$(\tau) \leftarrow \displaystyle\min_{\mu \in \text{CHILDREN}(\tau)}$ COST$(\mu)^a$
               CHILDREN$(\tau) \leftarrow \{\mu : \mu \in$ CHILDREN$(\tau)$ and COST$(\mu) =$ COST$(\tau)\}$

INSTANTIATIONS$(\tau)$
  01. when TERMS$(\tau)$ = NIL do
  02.     if COST$(\tau) = \infty$, TERMS$(\tau) \leftarrow \{\}$; return
  03.     for each $\mu \in$ CHILDREN$(\tau)$ do INSTANTIATIONS$(\mu)$
  04.     case LITERAL_NODE?$(\tau)$: TERMS$(\tau) \leftarrow \{\{$LITERAL_OF$(\tau)\}\}$
  05.         AND_NODE?$(\tau)$    : TERMS$(\tau) \leftarrow \displaystyle\bigcap_{\mu \in \text{CHILDREN}(\tau)}$ TERMS$(\mu)^b$
  06.         OR_NODE?$(\tau)$    : TERMS$(\tau) \leftarrow \displaystyle\bigcup_{\mu \in \text{CHILDREN}(\tau)}$ EXTEND$($TERMS$(\mu), Atoms(\tau, \mu))$

EXTEND$(\mathbf{T}, \mathbf{S})$
  01. if $\mathbf{S} = \{\}$
  02. then return $\mathbf{T}$
  03. else $V \leftarrow$ HEAD$(\mathbf{S})$
  04.     $\mathbf{Z} \leftarrow \{\{(V, v)\} : \mathcal{F}((V, v)) = 0\}$
  05.     EXTEND$(\mathbf{T} \cap \mathbf{Z},$ REST$(\mathbf{S}))$

---

a. min is defined with respect to the total order $\leq_\oplus$. When applied to an empty set, min returns the special symbol $\infty$ which is bigger than any cost in $\Sigma$.
b. $\{\alpha_1, \alpha_2, \ldots, \alpha_n\} \cap \{\beta_1, \beta_2, \ldots, \beta_m\} = \bigcup_{i=1}^{n} \bigcup_{j=1}^{m} \{\alpha_i \cup \beta_j\}$.

Figure 12: An algorithm for extracting minimal diagnoses from a decomposable NNF.

## Phase II: Computing Instantiations

The result of Phase I is a negation normal form that characterizes minimal diagnoses only. To extract the minimal diagnoses, all we need then is to compute the instantiations of this pruned negation normal form, that is, compute its sum-of-product representation. This can be done recursively as follows:

1. The instantiations of a literal $\mu$ are $\{\mu\}$.

2. The instantiations of a conjunction are the Cartesian product of the instantiations of its conjuncts.[17]

3. The instantiations of a disjunction are the union of the *extended* instantiations of its disjuncts. To understand this extension process, consider a disjunction $\mu$ and one of its disjuncts $\nu$. Just before the instantiations of $\nu$ are unioned with the instantiations of its siblings, they are extended by adding zero-cost literals to them. The goal of this extension process is to make sure that each extended instantiation of $\nu$ mentions all atoms that appear in $\mu$. This step would not be necessary if we were computing the DNF of $\mu$. Note, however, that are computing the sum-of-product representation of $\mu$ instead, which means that each product in the sum must mention all atoms that appear in $\mu$.

The soundness of the algorithm is given below:

**Theorem 13** *After the termination of* MINIMAL_INSTANTIATIONS *in Figure 12,* TERMS($\mu$) = *MinInst*($\mu$) *for every node $\mu$ in the NNF-graph $\tau$.*

A detailed example of the algorithm is given in Figure 13 where cost propagation, pruning, and the computation of instantiations are all explicated.

The computational complexity of the presented algorithm is given next:

**Theorem 14** *Consider the algorithm* MINIMAL_INSTANTIATIONS *in Figure 12. The time complexity of this algorithm is $O(cE)$ where $E$ is the number of edges in the NNF-graph $\tau$ and $c$ is square the number of instantiations returned by the algorithm (that is, $c = \mid MinInst(\tau) \mid^2$).*

We have two points to stress about the above result.

First, if the number of minimal diagnoses is small enough, the factor $c$ is a constant, and the time and space complexity of MINIMAL_INSTANTIATIONS becomes $O(E)$ which is linear in the size of $\tau$. However, if the number of minimal diagnoses cannot be regarded as a constant, then one cannot claim linear complexity. Needless to say, however, that one can never do better than the size of the answer that one is trying to compute. For example, if the number of minimal diagnoses is exponential in the size of $\tau$, then no extraction algorithm can do better than exponential since the answer it has to return is exponential.

Second, the size of the consequence generated by SYSTEM_CONSEQUENCE of Figure 8 is decided by the jointree $\mathcal{T}$ and the set of observed atoms $\mathbf{E}$ — it is independent of the

---

17. This step will not be sound unless the NNF is decomposable, that is, no atoms are shared between conjuncts.

## An Example of Computing Minimal Instantiations

Consider the consequence in Figure 9 and the minimal cardinality cost function. Calling PRUNE on the consequence leads to the following pruned NNF:

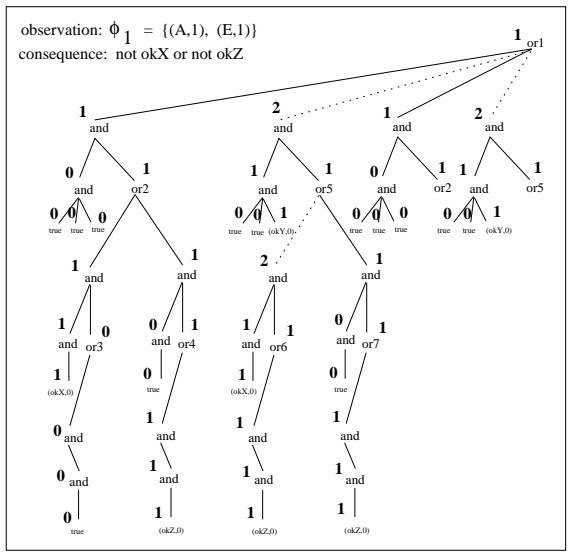

The cost of each node is shown next to it. The dotted links represent pruned parent-child links. Calling INSTANTIATIONS on this result leads to the following:

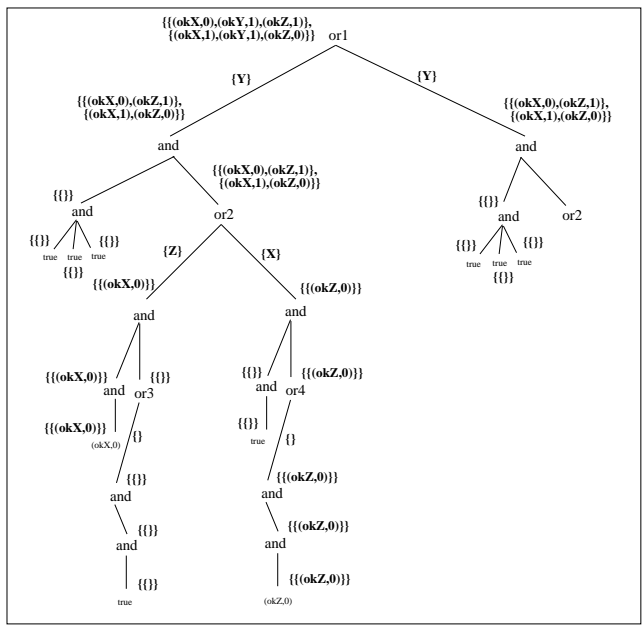

$Atoms(\mu, \nu)$ is shown on the arc between $\mu$ and $\nu$. TERMS$(\mu)$ is shown next to node $\mu$. The final result is two minimal instantiations $\neg okX \land okY \land okZ$ and $okX \land okY \land \neg okZ$.

Figure 13: An example illustrating the computation of minimal instantiations.

specific instantiation $\phi$ of $\mathbf{E}$.[18] However, the number of minimal diagnoses extracted by MINIMAL_INSTANTIATIONS in Figure 12 is very much dependent on the specific observation $\phi$. Therefore, if two observations $\phi_1$ and $\phi_2$ refer to the same set of atoms $\mathbf{E}$, the cost of generating their consequences will be the same. However, the cost of extracting their minimal diagnoses may be different.

For a concrete example of this last point, consider an $n$-bit adder and the following observations:

- $\phi_1$: All input/output bits are low except for the first sum bit.

- $\phi_2$: All input/output bits are low except for all sum bits.

The corresponding consequences of these observations will be both linear in $n$. However, the first observation leads to two minimal diagnoses — the first full adder has a broken xor-gate — while the second observation leads to $2^n$ minimal diagnoses — each of the full adders has a broken xor-gate. Therefore, computing the minimal diagnoses of $\phi_1$ is linear in $n$, but computing the minimal diagnoses of $\phi_2$ is exponential in $n$.

The last few observations reveal a merit of splitting the computation of minimal diagnoses into two steps: characterization of all diagnoses and then extraction of minimal ones. This is contrary to common practice in structure-based reasoning, but it allows one to make more refined statements on the complexity of model-based diagnosis.

## 7. Conclusion and Related Work

We have presented a comprehensive approach for characterizing and computing minimal diagnoses given a structured system description. What is most important about our approach is that it ties the computational complexity of diagnostic reasoning to a very meaningful aspect of systems: the topology of their system structures. Thus, it provides diagnostic practitioners with more flexibility in engineering the response time of their applications. This emphasis on structure has been the central theme in probabilistic reasoning lately and there are a number of other algorithms for importing this theme into model-based diagnosis (Dechter & Dechter, 1996; Geffner & Pearl, 1987). There are some key differences, however, between our proposal and the previous ones:

- Given consequences and the theorems to manipulate them, our proposal can be viewed as a framework for structure-based diagnosis instead of simply an algorithm. We did propose a specific algorithm which utilizes jointrees, but our framework can accommodate other structure-based algorithms (that do not necessarily utilize jointrees) as long as they provide a mechanism for applying the Intersection Corrollary efficiently (see (Darwiche, 1998b) for an example).

- We have decomposed the computation of minimal diagnoses into two independent stages: characterization of diagnoses using consequences and then choosing a minimal subset of them. This separation has at least three advantages. First, it allowed us to offer guarantees on the size of a consequence which are independent of the number of (minimal) diagnoses that it characterizes. Next, it allowed us to incorporate different minimality criteria without having to alter the characterization algorithm. Finally, it inspired a device compilation technique which is discussed in (Darwiche, 1998a).

---

18. Component consequences will depend on $\phi$, but not the number of and/or nodes which are added by SYSTEM_CONSEQUENCE.

- The time and space complexity of our algorithm depends not only on the system topology but also on the system observables. In particular, we have shown that the more observables we have, the easier it is to diagnose a system using our algorithm. We have also provided a refined complexity result which explicates precisely the effect of system observables on the complexity of structure-based diagnosis.

- Finally, previous algorithms have rested on the language of constraints among multi-valued variables. Our approach uses a purely logical setting, which allows computation directly on Boolean syntax. This bridges the gap even further between structure–based reasoning and model–based diagnosis.

## Acknowledgements

The author would like to thank Yousri El Fattah and Gregory Provan for providing useful comments on an earlier version of this paper.

## Appendix A. Representing Negation Normal Forms

This appendix describes our representation of negation normal forms using directed acyclic graphs. It also provides a number of operations on the suggested representation and some associated conventions. The material in this appendix is necessary to understand the pseudocode provided throughout the paper.

**Representation:**

An *NNF-graph* over atoms $\mathbf{V}$ is a rooted, directed acyclic graph with three types of nodes: literal-nodes, and-nodes and or-nodes. For each literal node $\mu$, LITERAL_OF$(\mu) = (V, v)$ where $V \in \mathbf{V}$ and $v \in \{0, 1\}$. Each node $\mu$ in an NNF-graph represents a Boolean sentence as follows:

- If $\mu$ is a literal-node and LITERAL_OF$(\mu) = (V, v)$, then $\mu$ represents
    - the positive literal $V$ when $v = 1$
    - the negative literal $\neg V$ when $v = 0$.
- If $\mu$ is an and-node and the $i^{th}$ child of $\mu$ represents sentence $\alpha_i$, then $\mu$ represents the conjunction $\wedge_i \alpha_i$.
- If $\mu$ is an or-node and the $i^{th}$ child of $\mu$ represents sentence $\alpha_i$, then $\mu$ represents the disjunction $\vee_i \alpha_i$.

The sentence represented by an NNF-graph is the sentence represented by its root node.

**Operations:**

We have the following operations to manipulate NNF-graphs:

- LITERAL_NODE?$(\mu)$, AND_NODE?$(\mu)$, OR_NODE?$(\mu)$: type predicates for nodes.
- NEW_LITERAL_NODE$(V, v)$: creates and returns a new literal-node labeled with $(V, v)$.
- NEW_AND_NODE(): creates and returns a new and-node (with no children).
- NEW_OR_NODE(): creates and returns a new or-node (with no children).
- ADD_CHILD$(\nu, \mu)$: adds node $\mu$ to the children of node $\nu$.
- CHILDREN$(\nu)$: returns the children of node $\nu$.

**Conventions:**

We adopt the following conventions with respect to NNF-graphs:

- If $\tau$ is an NNF-graph, then $\tau$ will also be used to denote the root of $\tau$.
- We use *true* to denote an and-node with no children and *false* to denote an or-node with no children.
- We do not show the directions of arcs assuming that they point downwards.
- The NNF-graph rooted at some node is formed from that node and all its descendants.
- A node and the sentence it represents will be used exchangeably.

**Example:**

Consider the following NNF-graph:

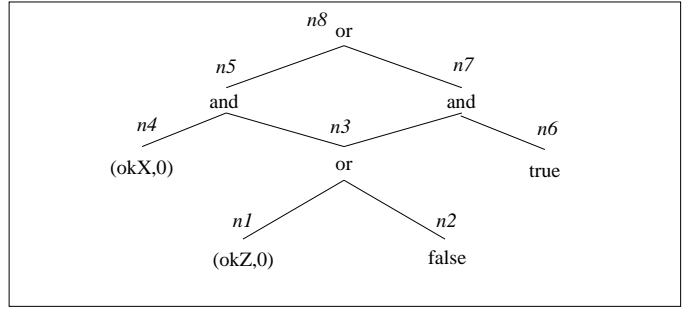

This graph represents the following sentence $(\neg okX \wedge (\neg okZ \vee false)) \vee ((\neg okZ \vee false) \wedge true)$. It can be created using the following sequence of calls:

$n_1 \leftarrow \text{NEW\_LITERAL\_NODE}(okZ, 0)$      $\text{ADD\_CHILD}(n_5, n_4)$

$n_2 \leftarrow \text{NEW\_OR\_NODE}()$      $n_6 \leftarrow \text{NEW\_AND\_NODE}()$

$n_3 \leftarrow \text{NEW\_OR\_NODE}()$      $n_7 \leftarrow \text{NEW\_AND\_NODE}()$

$\text{ADD\_CHILD}(n_3, n_1)$      $\text{ADD\_CHILD}(n_7, n_3)$

$\text{ADD\_CHILD}(n_3, n_2)$      $\text{ADD\_CHILD}(n_7, n_6)$

$n_4 \leftarrow \text{NEW\_LITERAL\_NODE}(okX, 0)$      $n_8 \leftarrow \text{NEW\_OR\_NODE}()$

$n_5 \leftarrow \text{NEW\_AND\_NODE}()$      $\text{ADD\_CHILD}(n_8, n_5)$

$\text{ADD\_CHILD}(n_5, n_3)$      $\text{ADD\_CHILD}(n_8, n_7)$.

## Appendix B. Representing Literals and Instantiations

This appendix discusses our representation of literals and instantiations and provides some operations on these representations. The material in this appendix is necessary to understanding the pseudocode which is provided throught the paper.

**Representation:**

An instantiation is represented by a set of pairs $(V, v)$ where $V$ is an atom and $v \in$ VALUES_OF$(V) = \{0, 1\}$.

**Operations:**

An instantiation $\mathbf{L}$ can be asserted or retracted using ASSERT$(\mathbf{L})$ and RETRACT$(\mathbf{L})$. After asserting an instantiation $\mathbf{L}$, we have the following for every pair $(V, v)$ in $\mathbf{L}$:

- INSTANTIATED?$(V)$ is true;
- VALUE_OF$(V)$ equals $v$.

We assume that sets of atoms are <u>ordered</u>, say alphabetically, where HEAD$(\mathbf{V})$ returns the first element of the set and REST$(\mathbf{V})$ returns the tail of the set. Two operations are defined on ordered sets of atoms:

- A function to compute a unique index for the instantiated values of set $\mathbf{V}$:[19]

    INDEX$(\mathbf{V})$ {*atoms in* $\mathbf{V}$ *must be instantiated*}
    01. if $\mathbf{V} = \{\}$
    02. then return 0
    03. else return VALUE_OF(HEAD$(\mathbf{V})$) $+ 2 *$ INDEX(REST$(\mathbf{V})$)

- A function to compute all instantiations of the uninstantiated atoms in $\mathbf{V}$:

    GENERATE_INSTANTIATIONS$(\mathbf{V})$ {*some atoms in* $\mathbf{V}$ *may be instantiated*}
    01. if $\mathbf{V} = \{\}$
    02. then return $\{\{\}\}$
    03. else $V \leftarrow$ HEAD$(\mathbf{V})$
    04.       $\mathbf{R} \leftarrow$ GENERATE_INSTANTIATIONS(REST$(\mathbf{V})$)
    05.       if INSTANTIATED?$(V)$ then return $\mathbf{R}$
    06.       else return $\bigcup\limits_{v \in \text{VALUES\_OF}(V)} \{\{(V, v)\} \cup \alpha : \ \alpha \in \mathbf{R}\}$

**Examples:**

Consider the instantiation $\alpha = A \wedge \neg B \wedge D$. This is represented as $\mathbf{L} = \{(A, 1), (B, 0), (D, 1)\}$. Suppose now that we call ASSERT$(\mathbf{L})$. Then INSTANTIATED?$(A)$ is true and VALUE_OF$(B) = 0$. Moreover, INDEX$(\{A, B, D\}) = 5$ and we have GENERATE_INSTANTIATIONS$(\{C, D, E\}) = \{\{(C, 1), (E, 1)\}, \{(C, 0), (E, 1)\}, \{(C, 1), (E, 0)\}, \ \{(C, 0), (E, 0)\}\}$.

---

19. The ordering of $\mathbf{V}$ is essential for this function to return the same index for each instantiation of $\mathbf{V}$.

## Appendix C. Avoiding the Sharing of Assumables in Component Descriptions

Consider the following figure which contains a structured system description where component descriptions $\Delta_C$ and $\Delta_D$ share the assumable $Pwr$:

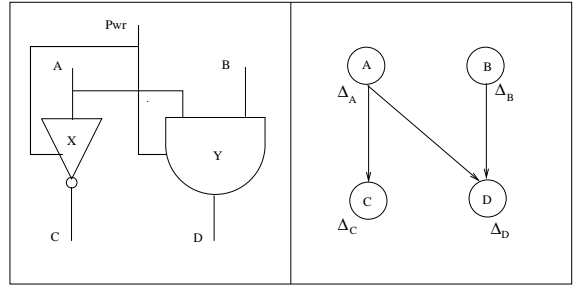

| $\Delta_A, \Delta_B$ | $\Delta_C$ | $\Delta_D$ |
|---|---|---|
| {} | $\left\{\begin{array}{rcl} A \wedge (Pwr \wedge okX) & \supset & \neg C \\ \neg A \wedge (Pwr \wedge okX) & \supset & C \\ \neg Pwr & \supset & \neg C \end{array}\right\}$ | $\left\{\begin{array}{rcl} A \wedge B \wedge (Pwr \wedge okY) & \supset & D \\ \neg(A \wedge B) \wedge (Pwr \wedge okY) & \supset & \neg D \\ \neg Pwr & \supset & \neg D \end{array}\right\}$ |

We can avoid the sharing of assumable $Pwr$ by introducing an auxiliary node $Pwr'$ and making it equivalent to the shared assumable $Pwr$ as shown in the figure below. As a result of adding this auxiliary node $Pwr'$, we no longer have a common assumable between the component descriptions $\Delta_C$ and $\Delta_D$ (but we have a common non-assumable $Pwr'$ instead).

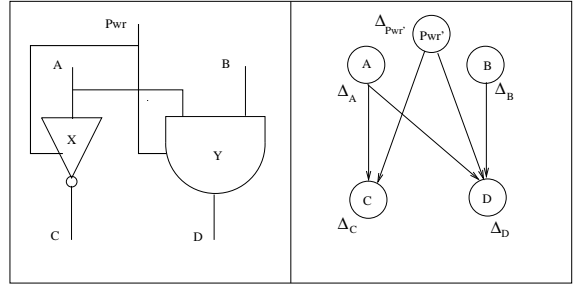

| $\Delta_A, \Delta_B$ | $\Delta_C$ | $\Delta_D$ | $\Delta_{Pwr'}$ |
|---|---|---|---|
| {} | $\left\{\begin{array}{rcl} A \wedge (Pwr' \wedge okX) & \supset & \neg C \\ \neg A \wedge (Pwr' \wedge okX) & \supset & C \\ \neg Pwr' & \supset & \neg C \end{array}\right\}$ | $\left\{\begin{array}{rcl} A \wedge B \wedge (Pwr' \wedge okY) & \supset & D \\ \neg(A \wedge B) \wedge (Pwr' \wedge okY) & \supset & \neg D \\ \neg Pwr' & \supset & \neg D \end{array}\right\}$ | $\left\{\begin{array}{rcl} Pwr & \supset & Pwr' \\ \neg Pwr & \supset & \neg Pwr' \end{array}\right\}$ |

This simple technique can be always used to ensure that no sharing of assumables takes place between component descriptions. Note, however, that this ensurance comes at a price: The structured system description with auxiliary nodes (no assumable sharing) will typically be topologically more complex than the one where sharing is allowed. This is clearly the case in the above example where the no-sharing-of-assumables led us to transform the system structure from a tree to a graph.

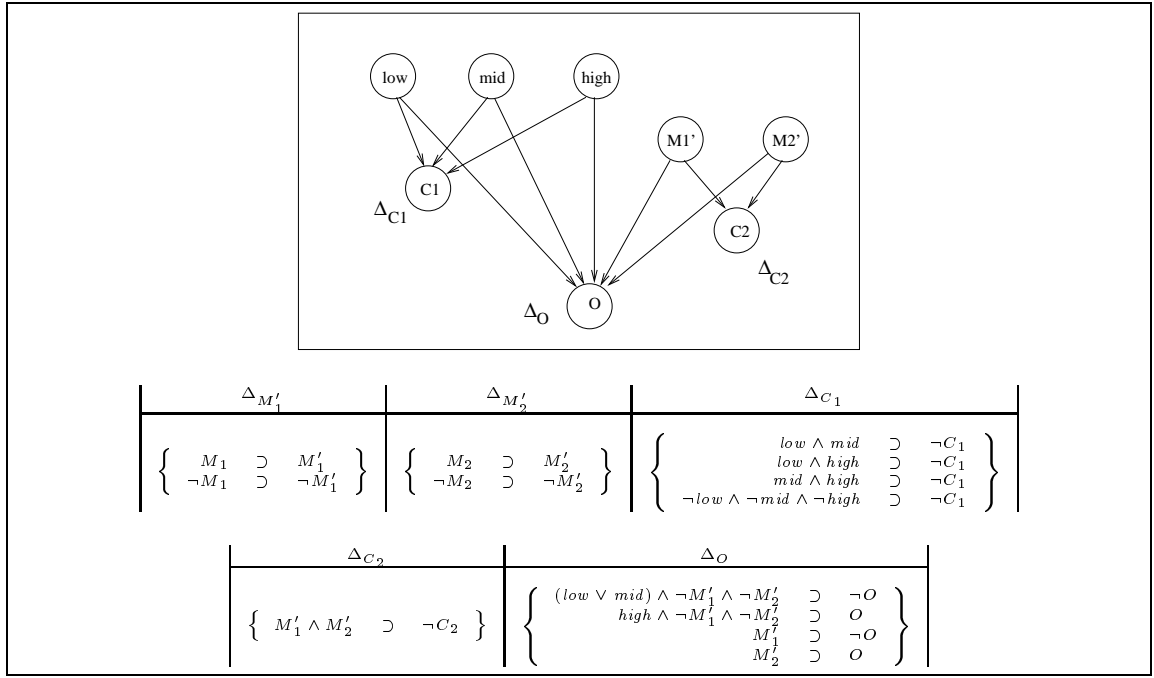

Figure 14: Modeling a component with multiple fault modes and multiple input values.

## Appendix D. Extending the Framework to Multivalued Variables

We show in this appendix how to represent constraints among assumables using two methods:

1. introducing auxiliary atoms into the system structure;

2. using multivalued variables instead of atoms.

The two approaches will be illustrated using the following example.

Consider a component which has one input and one output: the input is either *low*, *mid* or *high* and the output is either *on* or *off*. Furthermore, suppose that this component has two fault modes $M_1$ and $M_2$. Such a component can be represented as shown in Figure 14. In this figure, we have two auxiliary atoms, $C_1$ and $C_2$, which represent constraints. The first constraint, $C_1$, ensures that *low*, *mid* and *high* are mutually exclusive and exhaustive. The second constraint, $C_2$, ensures that $M_1$ and $M_2$ are mutually exclusive. The atoms $C_1$ and $C_2$ should always be observed to enforce these constraints. That is, for any system observation $\phi$, we must actually invoke SYSTEM_CONSEQUENCE on $\phi \wedge C_1 \wedge C_2$ to make sure that the two constraints are enforced.

This approach is relatively reasonable but the representation is not as concise as one would expect. A better solution for representing such systems involves the use of *multivalued variables* instead of atoms. This solution will be described next and it leads to a more concise, efficient representation.

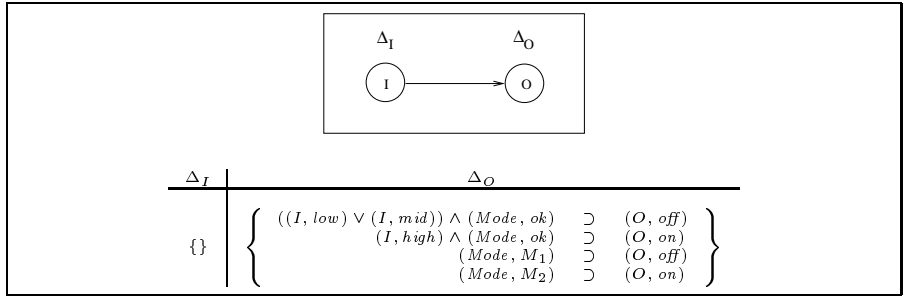

Figure 15: An SSD using multivalued variables.

An atom is a special case of a variable in that it only has two values. A general variable, however, can have any finite number of values. In the previous example, it would be best to represent the input as a variable with three values *low*, *mid* and *high*; to represent the output as a variable with two values *on* and *off*; and to represent the mode as a variable with three values $ok, M_1$ and $M_2$.

Using variables instead of atoms, the system in Figure 14 can be encoded as given in Figure 15. This is clearly a more compact and efficient representation.

More generally though, we extend our framework to variables by taking the set of non-assumables **P** and the set of assumables **A** to be sets of variables, instead of atoms. We next discuss formally the syntax and semantics of the resulting variable-based logic, and then show what modifications are needed to our presented definitions and pseudocode (only two changes are needed).

**Syntax and Semantics** In this variable-based framework, a literal is a pair $(V, v)$, where $V$ is a variable and $v$ is a value. A sentence is either a literal, the negation of a sentence, or the combination of two sentences using a standard logical connective such as $\land$, $\lor$ or $\supset$. This defines the syntax of a propositional logic with multivalued variables. This logic includes standard propositional logic as a special case (that is, when all variables are binary). The semantics of this logic is defined in the usual way. That is, a model $\omega$ is a function that assigns one value to each variable. Moreover, a model $\omega$ satisfies a sentence under the following conditions:

$\omega \models (V, v)$ if $\omega$ assigns the value $v$ to variable $V$;

$\omega \models \neg \alpha$ if $\omega \not\models \alpha$;

$\omega \models \alpha \lor \beta$ if $\omega \models \alpha$ or $\omega \models \beta$; and

$\omega \models \alpha \land \beta$ if $\omega \models \alpha$ and $\omega \models \beta$.

A sentence is satisfiable (consistent) iff it has at least one model. This also leads to the usual notions of entailment and validity.

It is important to stress that in this variable-based logic, the term $\neg(V, v)$ is not considered a literal in general. Note that if $V$ is not a binary variable, then $\neg(V, v)$ would be equivalent to a clause $(V, v_1) \lor (V, v_2) \lor \ldots \lor (V, v_n)$ where $v_1, \ldots, v_n$ are the values of variable $V$ not equal to $v$. Having said that, an instantiation of a set of variables **V** is defined as a conjunction of literals, one literal for each variable in $\mathbf{V}$. This means that $\neg(I, low) \vee (O, on) \vee (M, ok)$ is not an instantiation according to the previous definition.

**Modifications**     In the extended framework, all definitions and theorems remain intact with two exceptions:

1. Since variables can have more than two values, we need to generalize the function INDEX in Appendix B which computes unique indices for instantiations:

   INDEX($\mathbf{V}$)
      01. $(i, b) \leftarrow$ AUX_INDEX($\mathbf{V}$)
      02. return $i$

   AUX-INDEX($\mathbf{V}$)
      01. if $\mathbf{V} = \{\}$
      02. then return $(0, 1)$
      03. else $(i, b) \leftarrow$ AUX_INDEX(REST($\mathbf{V}$))
      04.    $V \leftarrow$ HEAD($\mathbf{V}$)
      05.    return $(i + \text{VALUE\_OF}(V) * b, | \text{VALUES\_OF}(V) | * b)$

   Here, we are assuming that the values of each variable are $\{0, 1, \ldots\}$. Suppose, for example, that we have $\mathbf{V} = \{V_1, V_2, V_3\}$ where $V_1$ and $V_3$ have the values $\{0, 1\}$ and $V_2$ has the values $\{0, 1, 2\}$. We then have the following instantiations and their corresponding indices:

   | $V_1$ | $V_2$ | $V_3$ | index |
   |---|---|---|---|
   | 0 | 0 | 0 | 0 |
   | 0 | 0 | 1 | 1 |
   | 0 | 1 | 0 | 2 |
   | 0 | 1 | 1 | 3 |
   | 0 | 2 | 0 | 4 |
   | 0 | 2 | 1 | 5 |
   | 1 | 0 | 0 | 6 |
   | 1 | 0 | 1 | 7 |
   | 1 | 1 | 0 | 8 |
   | 1 | 1 | 1 | 9 |
   | 1 | 2 | 0 | 10 |
   | 1 | 2 | 1 | 11 |

2. Since assumables can have more than two values, Definition 12 of a cost function should be changed so that the condition:

   - $\mathcal{F}$ maps each $\mathbf{A}$-literal into a cost in $\Sigma$, where each literal or its negation has cost 0;

   reads as

   - $\mathcal{F}$ maps each $\mathbf{A}$-literal into a cost in $\Sigma$, where for each variable $V \in \mathbf{A}$, at least one literal $(V, v)$ must have cost 0.

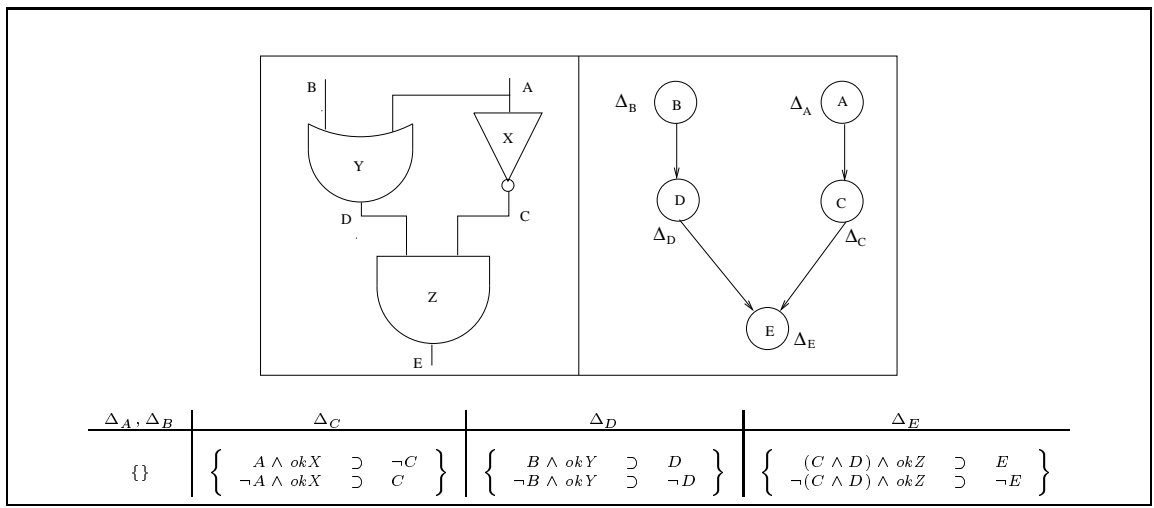

| $\Delta_A, \Delta_B$ | $\Delta_C$ | $\Delta_D$ | $\Delta_E$ |
|---|---|---|---|
| $\{\}$ | $\left\{\begin{array}{lll} A \wedge okX & \supset & \neg C \\ \neg A \wedge okX & \supset & C \end{array}\right\}$ | $\left\{\begin{array}{lll} B \wedge okY & \supset & D \\ \neg B \wedge okY & \supset & \neg D \end{array}\right\}$ | $\left\{\begin{array}{lll} (C \wedge D) \wedge okZ & \supset & E \\ \neg(C \wedge D) \wedge okZ & \supset & \neg E \end{array}\right\}$ |

Figure 16: A structured system description of a digital circuit. The arc from node $A$ to node $D$ has been cut out given that node $A$ is observed.

## Appendix E. Cutting Outgoing-Arcs of Observed Nodes

Consider the structured system description given in Figure 3 and suppose that we want to compute the diagnoses for system observation $\phi = \neg A \wedge \neg E$. This computation can be performed with respect to the SSD in Figure 3 or, equivalently, it can be performed with respect to the SSD in Figure 16. This SSD is obtained from the one in Figure 3 as follows:

1. The arc going from node $A$ to node $D$ is cut out.

2. Every occurrence of atom $A$ in the component description $\Delta_D$ is replaced with *false* since $A$ appears negated in $\phi$.[20]

It is easy to show that the consequence of $\phi$ is the same with respect to both SSDs. Working with the modified SSD is preferred, however, since it has a tree system structure.

We have the following observations about this process of cutting out arcs from the system structure:

- One can cut out all outgoing-arcs of any node that appears in the system observation. In our previous example, we can also cut out the arc from $A$ to $C$ but we must also modify $\Delta_C$ accordingly.[21]

- Even if cutting out arcs does not lead to a tree structure, it would typically lead to reducing the size of cliques of the resulting system structure. In fact, the reduction could move some problems from being practically unsolvable using structure-based methods to being solvable. Consider the results reported in (El Fattah & Dechter,

---

20. $A$ would be replaced by *true* if it did not appear negated.
21. This will lead to a disconnected system structure. In such a case, we have to compute the consequence of each disconnected piece and then conjoin all consequences.

1996) for an example which show jointree statistics for the benchmark circuits proposed in (Beglez & Fujiwara, 1985). The following table shows the maximal-clique sizes reported for a few of the circuits together with the maximal-clique sizes after the arcs outgoing from observed nodes (root nodes) are cut out:[22]

| Circuit | #nodes | maximal-clique size before cutting arcs | maximal-clique size after cutting arcs |
|---------|--------|------------------------------------------|-----------------------------------------|
| c432 | 196 | 28 | 22 |
| c499 | 243 | 25 | 10 |
| c880 | 443 | 28 | 10 |
| c1355 | 587 | 25 | 10 |

There is clearly a dramatic change in the maximal-clique size for the last three circuits. In fact, there are only a few cliques which have the maximal size or a size close to it as is shown in (El Fattah & Dechter, 1996). This makes the approach reported in this paper appropriate for the last three circuits. The approach, however, is not appropriate as is for the first circuit.

We close this section by stressing again that this process of cutting out arcs can lead to a very dramatic reduction in maximal-clique size. Therefore, it should be exploited whenever possible (Darwiche & Provan, 1997).

---

22. The jointrees after cutting out arcs are computed using the algorithm reported in (Huang & Darwiche, 1996).

## Appendix F. Proofs

### Proof of Theorem 1

Suppose that $Cons_{\mathbf{A}}^{\Delta}(\phi) \equiv \alpha_1 \vee \ldots \vee \alpha_n$ where each $\alpha_i$ is an **A**-instantiation. Then $\Delta \cup \{\phi\} \models \alpha_1 \vee \ldots \vee \alpha_n$ by definition of a consequence. We need to prove two things:

1. Each $\alpha_i$ is a diagnosis.
   It suffices to show that $\Delta \cup \{\phi, \alpha_i\}$ is consistent. Suppose that $\Delta \cup \{\phi, \alpha_i\}$ is not consistent. Then $\Delta \cup \{\phi\} \models \neg\alpha_i$ and

$$\Delta \cup \{\phi\} \models \alpha_1 \vee \ldots \vee \alpha_{i-1} \vee \alpha_{i+1} \vee \ldots \vee \alpha_n.$$

   Taking $\beta = \alpha_1 \vee \ldots \vee \alpha_n$, we contradict Condition 3 in the definition of a consequence.

2. If $\beta$ is a diagnosis, then $\beta$ must be equal to some $\alpha_i$.
   Suppose that $\beta$ is a diagnosis. Then $\Delta \cup \{\phi, \beta\}$ must be consistent. Moreover, $\Delta \cup \{\phi, \beta\} \models \alpha_1 \vee \ldots \vee \alpha_n$ since $\Delta \cup \{\phi\} \models \alpha_1 \vee \ldots \vee \alpha_n$. This means that $\beta$ must be equal to some $\alpha_i$; otherwise, $\Delta \cup \{\phi, \beta\}$ would be inconsistent which we know it is not. □

### Proof of Theorem 2

Follows directly from the definition of a consequence and the characterization theorem. □

### Proof of Theorem 3

First, observe that each clause $\alpha$ in $\Delta_O$ can be written as the disjunction of two clauses $\alpha_{\mathbf{P}} \vee \alpha_{\mathbf{A}}$ where $\alpha_{\mathbf{P}}$ is the projection of $\alpha$ on non-assumables **P** and $\alpha_{\mathbf{A}}$ is the projection of $\alpha$ on assumables **A**. Moreover, $\phi$ mentions every atom that appears in $\alpha_{\mathbf{P}}$. Therefore, either

1. $\phi \models \alpha_{\mathbf{P}}$: Hence, $\phi \models \alpha_{\mathbf{P}} \vee \alpha_{\mathbf{A}}$ and $\phi \wedge (\alpha_{\mathbf{P}} \vee \alpha_{\mathbf{A}})$ is equivalent to $\phi$; or

2. $\phi \models \neg\alpha_{\mathbf{P}}$: Hence, $\phi \wedge (\alpha_{\mathbf{P}} \vee \alpha_{\mathbf{A}})$ is equivalent to $\phi \wedge \alpha_{\mathbf{A}}$.

Therefore, $\Delta_O \cup \{\phi\}$ is equivalent to $\phi$ conjoined with every $\alpha_{\mathbf{A}}$ whose matching $\alpha_{\mathbf{P}}$ is inconsistent with $\phi$. Therefore, $Cons_{\mathbf{A}}^{\Delta_O}(\phi)$ is the conjunction of all $\alpha_{\mathbf{A}}$'s whose matching $\alpha_{\mathbf{P}}$ is inconsistent with $\phi$. □

### Proof of Theorem 4

We prove this theorem in two steps:

1. $Cons^{\Theta \cup \Gamma}(\phi) \equiv Cons^{\Theta}(\phi) \wedge Cons^{\Gamma}(\phi)$.
   Recall that $Cons^{\Theta \cup \Gamma}(\phi)$ is the strongest **A**-sentence entailed by $\Theta \cup \Gamma \cup \{\phi\}$. By Lemma 3:

   - There is a database $\Theta'$ that does not share atoms with $\phi$, and $\Theta' \cup \{\phi\} \equiv \Theta \cup \{\phi\}$.

   - There is a database $\Gamma'$ that does not share atoms with $\phi$, and $\Gamma' \cup \{\phi\} \equiv \Gamma \cup \{\phi\}$.

Note that no clause in $\Theta' \cup \{\phi\}$ can resolve with any clause in $\Gamma' \cup \{\phi\}$ since there are no common atoms between $\Theta'$ and $\Gamma'$. Therefore, the **A**-clauses derived from $\Theta' \cup \Gamma' \cup \{\phi\}$ are those derived from $\Theta' \cup \{\phi\}$ in addition to those derived from $\Gamma' \cup \{\phi\}$.[23] It then follows that the strongest **A**-sentence entailed by $\Theta' \cup \Gamma' \cup \{\phi\}$ — which is equivalent to the conjunction of all **A**-clauses entailed by $\Theta' \cup \Gamma' \cup \{\Phi\}$ — is equivalent to the strongest **A**-sentence entailed by $\Theta' \cup \{\phi\}$ conjoined with the strongest **A**-sentence entailed by $\Gamma' \cup \{\phi\}$:

$$Cons^{\Theta' \cup \Gamma'}(\phi) \equiv Cons^{\Theta'}(\phi) \wedge Cons^{\Gamma'}(\phi),$$

which leads to:

$$Cons^{\Theta \cup \Gamma}(\phi) \equiv Cons^{\Theta}(\phi) \wedge Cons^{\Gamma}(\phi).$$

2. $Cons^{\Theta}(\phi) \equiv Cons^{\Theta}(\phi_{\Theta})$ and $Cons^{\Gamma}(\phi) \equiv Cons^{\Gamma}(\phi_{\Gamma})$.

   The atoms that appear in $\phi$ but do not appear in $\phi_{\Theta}$ do not appear in $\Theta$ either. Therefore, they do not affect the strongest **A**-sentence entails by $\Theta \cup \{\phi\}$. The same is true for $\Gamma$ and $\phi_{\Gamma}$.

Therefore,

$$Cons^{\Theta \cup \Gamma}(\phi) \equiv Cons^{\Theta}(\phi_{\Theta}) \wedge Cons^{\Gamma}(\phi_{\Gamma}). \quad \square$$

**Lemma 3** *Suppose that $\Delta$ is a database and $\phi$ is a set of literals. There exists a database $\Delta'$ such that*

1. *$\Delta'$ does not share atoms with $\phi$; and*

2. *$\Delta' \cup \phi$ is equivalent to $\Delta \cup \phi$.*

**Proof of Lemma 3**

We show how to construct $\Delta'$ from $\Delta$. Suppose that $\Delta$ is in clausal form. Each clause $\alpha$ in $\Delta$ must satisfy one of the following conditions:

1. $\alpha$ does not share atoms with $\phi$;

2. $\alpha$ shares some atoms with $\phi$ and we have either:

   (a) $\alpha$ and $\phi$ share a literal, which means that $\alpha$ is subsumed by $\phi$; or

   (b) $\alpha$ and $\phi$ share no literals, which means that $\alpha$ resolves with some literals in $\phi$ to yield a clause $\beta$ that does not share atoms with $\phi$. Moreover, $\phi \cup \{\beta\}$ is equivalent to $\phi \cup \{\alpha\}$.

We can obtain $\Delta'$ from $\Delta$ as follows. For each clause $\alpha$ in $\Delta$:

1. if $\alpha$ is in Class 1 above, add $\alpha$ to $\Delta'$;

2. if $\alpha$ is in Class 2a above, ignore $\alpha$;

3. if $\alpha$ is in Class 2b above, add the resolvant $\beta$ to $\Delta'$.

It should be obvious that $\Delta' \cup \phi$ is equivalent to $\Delta \cup \phi$. $\quad \square$

---

23. An **A**-clause is a clause which contains **A**-literals only.

**Proof of Theorem 5**

It suffices to prove that $Cons^\Delta(\gamma \vee \beta) \equiv Cons^\Delta(\gamma) \vee Cons^\Delta(\beta)$. The theorem then follows from observing that $\phi \equiv \bigvee_\alpha (\phi \wedge \alpha)$ where $\alpha$ ranges over instantiations of atoms $\mathbf{S}$ that are consistent with $\phi$. This leads to $Cons^\Delta(\phi) \equiv Cons^\Delta(\bigvee_\alpha (\phi \wedge \alpha)) \equiv \bigvee_\alpha Cons^\Delta(\phi \wedge \alpha)$.

1. $Cons^\Delta(\gamma) \vee Cons^\Delta(\beta) \models Cons^\Delta(\gamma \vee \beta)$.
   By definition of a consequence, we have $Cons^\Delta(\gamma) \models Cons^\Delta(\gamma \vee \beta)$ and $Cons^\Delta(\beta) \models Cons^\Delta(\gamma \vee \beta)$. Therefore, we must have $Cons^\Delta(\gamma) \vee Cons^\Delta(\beta) \models Cons^\Delta(\gamma \vee \beta)$.

2. $Cons^\Delta(\gamma \vee \beta) \models Cons^\Delta(\gamma) \vee Cons^\Delta(\beta)$.
   By definition of a consequence, we have $\Delta \cup \{\gamma\} \models Cons^\Delta(\gamma)$ and $\Delta \cup \{\beta\} \models Cons^\Delta(\beta)$. Therefore, $(\Delta \cup \{\gamma\}) \vee (\Delta \cup \{\beta\}) \models Cons^\Delta(\gamma) \vee Cons^\Delta(\beta)$. This is equivalent to $\Delta \cup \{\gamma \vee \beta\} \models Cons^\Delta(\gamma) \vee Cons^\Delta(\beta)$, which means that $Cons^\Delta(\gamma) \vee Cons^\Delta(\beta)$ is an $\mathbf{A}$-sentence implied by $\Delta \cup \{\gamma \vee \beta\}$. By definition of $Cons^\Delta(\gamma \vee \beta)$, we must then have $Cons^\Delta(\gamma \vee \beta) \models Cons^\Delta(\gamma) \vee Cons^\Delta(\beta)$. □

**Proof of Theorem 6**

Suppose that $(\mathbf{P}, \mathbf{A}, \mathcal{G}, \Delta)$ is a structured system description according to Definition 11. We want to prove that $\Delta \cup \{\alpha\}$ is consistent for any instantiation $\alpha$ of the assumables $\mathbf{A}$.

The proof is by induction on the system structure.

**Base case:** Suppose that the system structure has a single node $O$. Then $\Delta = \Delta_O$ and it follows from the definition of $\Delta_O$ that $\Delta \cup \{\alpha\}$ is consistent for any instantiation $\alpha$ of the assumables $\mathbf{A}$.

**Inductive step:** Suppose that we have a structured system description $(\mathbf{P}, \mathbf{A}, \mathcal{G}, \Delta)$ satisfying the above property. It suffices to show that the property will still hold after we add a leaf node $O$ to the system structure together with its component description $(\mathbf{I}, O, \Delta_O)$. That is, we need to show that $\Delta \cup \Delta_O \cup \{\alpha\}$ is consistent for any instantiation $\alpha$ of the assumables $\mathbf{A}$ given that $\Delta \cup \{\alpha\}$ is consistent.

It suffices to show that $\Delta \cup \Delta_O \cup \{\alpha, \beta\}$ is consistent for some instantiation $\beta$ of $\mathbf{I} \cup \{O\}$. Since $\Delta \cup \{\alpha\}$ is consistent, there must exist some instantiation $\beta_\mathbf{I}$ of $\mathbf{I}$ such that $\Delta \cup \{\alpha, \beta_\mathbf{I}\}$ is consistent. Moreover, since $\Delta_O \cup \{\alpha, \beta_\mathbf{I}\}$ is consistent (by definition of a component description), there must exist some instantiation $\beta_O$ of $O$ such that $\Delta_O \cup \{\alpha, \beta_\mathbf{I}, \beta_O\}$ is consistent. Consider now the following:

- $\Delta_O \cup \{\alpha, \beta_\mathbf{I}, \beta_O\}$ is equivalent to $\{\alpha, \beta_\mathbf{I}, \beta_O\}$ since every atom that appears in $\Delta_O$ appears in the instantiation $\{\alpha, \beta_\mathbf{I}, \beta_O\}$ and, hence, $\{\alpha, \beta_\mathbf{I}, \beta_O\} \models \Delta_O$.

- $\Delta \cup \{\alpha, \beta_\mathbf{I}, \beta_O\}$ is consistent since $\Delta \cup \{\alpha, \beta_\mathbf{I}\}$ is consistent and the atom $O$ does not appear in $\Delta \cup \{\alpha, \beta_\mathbf{I}\}$.

Therefore, $\Delta \cup \Delta_O \cup \{\alpha, \beta_\mathbf{I}, \beta_O\} \equiv \Delta \cup \{\alpha, \beta_\mathbf{I}, \beta_O\}$ is consistent and, hence, $\Delta \cup \Delta_O \cup \{\alpha\}$ is consistent. □

**Proof of Theorem 7**

First, the call EMPTY?$(\alpha_{\mathbf{P}} \cap \gamma)$ on line 06 returns true iff no literal in $\alpha_{\mathbf{P}}$ appears in $\gamma$. Since $\alpha_{\mathbf{P}}$ represents a clause of some atoms in $\mathbf{I} \cup \{O\}$ and $\gamma$ represents an instantiation of $\mathbf{I} \cup \{O\}$, then every literal in $\alpha_{\mathbf{P}}$ is contradicted by some literal in $\gamma$. Therefore, this call will return true iff $\gamma \models \neg\alpha_{\mathbf{P}}$.

Second, the call CREAT_DISJUNCTION$(\alpha_{\mathbf{A}})$ on line 07 returns a disjunction of the literals in $\alpha_{\mathbf{A}}$.

Therefore, for each instantiation $\gamma$ of $\mathbf{I} \cup \{O\}$, the function COMPONENT_CONSEQUENCE computes a conjunction of disjunctions where each disjunction represents an $\alpha_{\mathbf{A}}$ whose matching $\alpha_{\mathbf{P}}$ is inconsistent with $\gamma$. The algorithm is then sound given Theorem 3. $\square$

**Proof of Theorem 8**

Line 01 will generate $2^{|\mathbf{I}|+1}$ instantiations, which is the number of times that lines 02-08 will repeat. Each of lines 02, 03 and 08 take $O(n)$ time. Line 04 takes $O(1)$ time. The loop in lines 05-07 will repeat $s$ times. Line 06 takes $O(n)$ time since EMPTY? can be implemented in $O(n)$ time. Line 07 also takes $O(n)$ time since CREAT_DISJUNCTION takes $O(n)$ time. Therefore, lines 05-07 take $O(sn)$ time and, hence, lines 02-08 take $O(sn)$. Finally, lines 01-09 take $2^{|\mathbf{I}|+1}O(sn)$ time which is $O(sn2^{|\mathbf{I}|})$. The space complexity is no worse than the time complexity since adding a node or arc to the computed consequence takes $O(1)$ time. $\square$

**Proof of Lemma 1**

First, we need to prove that atoms $\mathcal{C}_i$ are instantiated to some $\alpha$ when the call CLIQUE_CONSEQUENCE$(\mathcal{C}_i)$ is made. This follows immediately since CLIQUE_CONSEQUENCE$(\mathcal{C}_i)$ is only called on line 06 of SUBTREE_CONSEQUENCE and $\alpha$ is instantiated on line 04 of the same function.

The call CLIQUE_CONSEQUENCE$(\mathcal{C}_i)$ computes the following expression:

$$\bigwedge_{O \in \text{COMPONENTS\_OF}(\mathcal{C}_i)} \text{CONSEQUENCES}_O[\text{INDEX}(O \cup \mathcal{G}_O)].$$

Therefore, all we need to show is that

$$Cons_{\mathbf{A}}^{\Delta_i}(\alpha) \equiv \bigwedge_{O \in \text{COMPONENTS\_OF}(\mathcal{C}_i)} \text{CONSEQUENCES}_O[\text{INDEX}(O \cup \mathcal{G}_O)].$$

Given Theorem 7, it is enough to show that

$$Cons_{\mathbf{A}}^{\Delta_i}(\alpha) \equiv \bigwedge_{O \in \text{COMPONENTS\_OF}(\mathcal{C}_i)} Cons^{\Delta_O}(\alpha_{\Delta_O}),$$

where $(O, \mathcal{G}_O, \Delta_O)$ is the component description of $O$ and $\alpha_{\Delta_O}$ is the projection of instantiation $\alpha$ on the atoms $\{O\} \cup \mathcal{G}_O$ appearing in $\Delta_O$.

First note that $\alpha$ is an instantiation of the atoms in clique $\mathcal{C}_i$. Therefore, the common atoms between any two $\Delta_O$'s must appear in $\alpha$ and, hence, the Decomposition Theorem

gives:

$$
\begin{aligned}
Cons_{\mathbf{A}}^{\Delta_i}(\alpha) &\equiv Cons_{\mathbf{A}}^{\bigcup_{O \in \text{COMPONENTS\_OF}(\mathcal{C}_i)} \Delta_O}(\alpha) \\
&\equiv \bigwedge_{O \in \text{COMPONENTS\_OF}(\mathcal{C}_i)} Cons_{\mathbf{A}}^{\Delta_O}(\alpha_{\Delta_O}). \ \square
\end{aligned}
$$

## Proof of Lemma 2

We start with some concrete examples of the used notation. In Figure 5, we have:

$$
\Delta_{12} = \Delta_A \cup \Delta_B \cup \Delta_D = \left\{ \begin{array}{rcl} (A \vee B) \wedge ok\,Y & \supset & D, \\ \neg(A \vee B) \wedge ok\,Y & \supset & \neg D \end{array} \right\};
$$

while

$$
\Delta_{21} = \Delta_C \cup \Delta_E = \left\{ \begin{array}{rcl} A \wedge okX & \supset & \neg C, \\ \neg A \wedge okX & \supset & C, \\ C \wedge D \wedge okZ & \supset & E, \\ \neg(C \wedge D) \wedge okZ & \supset & \neg E \end{array} \right\}.
$$

Moreover, given that $\phi = A \wedge E$, then $\phi_{12} = A$ and $\phi_{21} = A \wedge E$.

SOUNDNESS OF CACHING

We will prove the lemma ignoring the caching on lines 00, 01 and 11. The soundness of this caching process follows because according to the lemma, any two calls to SUBTREE_CONSEQUENCE($\mathcal{C}_i, \mathcal{C}_j$) will return equivalent sentences if they are made under the same instantiation $\alpha$ of sepset $\mathcal{S}_{ij}$. Line 00 computes a unique index $l$ for each instantiation of $\mathcal{S}_{ij}$ which is used on line 01 to check if a previous call with respect to this instantiation has been made. If not, the computed sentence is cached on line 11 under the index $l$.

PROOF BY INDUCTION

First, we need to prove that atoms $\mathcal{S}_{ij}$ are instantiated when the call SUBTREE_CONSEQUENCE($\mathcal{C}_i, \mathcal{C}_j$) is made. SUBTREE_CONSEQUENCE is only called in two places:

1. SUBTREE_CONSEQUENCE($\mathcal{C}_z, \mathcal{C}_0$) is called on line 03 of SYSTEM_CONSEQUENCE: the sepset $\mathcal{S}_{z0} = \emptyset$ which is trivially instantiated.

2. SUBTREE_CONSEQUENCE($\mathcal{C}_k, \mathcal{C}_i$) is called on line 08 of SUBTREE_CONSEQUENCE: the sepset $\mathcal{S}_{ki}$ is a subset of clique $\mathcal{C}_i$ and the atoms of this clique are instantiated on line 04. Therefore, the atoms of $\mathcal{S}_{ki}$ are instantiated when this call to SUBTREE_CONSEQUENCE is made.

The rest of the proof is by induction on the structure of the jointree.

**Base case:** $\mathcal{C}_i$ has a single neighbor $\mathcal{C}_j$.

In this case, lines 08 of SUBTREE_CONSEQUENCE will not be executed. Therefore, the function SUBTREE_CONSEQUENCE is only computing the disjunction of all calls to CLIQUE_CONSEQUENCE($\mathcal{C}_i$) where each such call is made with respect to some instantiation $\alpha$ of clique $\mathcal{C}_i$ which is consistent with $\beta \wedge \phi_i$.

Given Lemma 1, all we need to show then is that:

$$Cons^{\Delta_{ij}}(\beta \wedge \phi_{ij}) \equiv \bigvee_\alpha Cons^{\Delta_i}(\alpha \wedge \phi_i).$$

By the Case–Analysis Theorem, we have:

$$Cons^{\Delta_{ij}}(\beta \wedge \phi_{ij}) \equiv \bigvee_\alpha Cons^{\Delta_{ij}}(\alpha \wedge \beta \wedge \phi_{ij}),$$

where $\alpha$ ranges over all instantiations of clique $\mathcal{C}_i$ which are consistent with $\beta \wedge \phi_{ij}$. We also have

- $\Delta_{ij}$ equals $\Delta_i$ and $\phi_{ij} = \phi_i$ since the subtree $\mathcal{T}_{ij}$ contains only the clique $\mathcal{C}_i$.

- $\alpha \wedge \beta \wedge \phi_{ij}$ is equivalent to $\alpha \wedge \beta \wedge \phi_i$.

- $\alpha \wedge \beta \wedge \phi_i$ is equivalent to $\alpha \wedge \phi_i$ since $\alpha \wedge \beta \wedge \phi_i$ is consistent and the atoms in $\beta$ are a subset of those in $\alpha \wedge \phi_i$.

Therefore,

$$Cons^{\Delta_{ij}}(\beta \wedge \phi_{ij}) \equiv \bigvee_\alpha Cons^{\Delta_i}(\alpha \wedge \phi_i),$$

which is what we need to show.

**Inductive step:**  Clique $\mathcal{C}_i$ has more than one neighbor.

The induction hypothesis is that the call SUBTREE_CONSEQUENCE($\mathcal{C}_k, \mathcal{C}_i$) on line 08 will return a sentence equivalent to $Cons^{\Delta_{ki}}(\alpha_{ki} \wedge \phi_{ki})$ where $\alpha_{ki}$ is the instantiation of sepset $\mathcal{S}_{ki}$.

Given this induction hypothesis, we want to show that the call SUBTREE_CONSEQUENCE($\mathcal{C}_i, \mathcal{C}_j$) will return a sentence equivalent to $Cons^{\Delta_{ij}}(\beta \wedge \phi_{ij})$ where $\beta$ is the instantiation of sepset $\mathcal{S}_{ij}$ before the call is made.

Given the induction hypothesis and Lemma 1, lines 02-10 are setting $Disj$ to the following expression

$$\bigvee_\alpha Cons^{\Delta_i}(\alpha) \wedge \bigwedge_{k \neq j} Cons^{\Delta_{ki}}(\alpha_{ki} \wedge \phi_{ki})$$

where $\alpha$ is the instantiation of clique $\mathcal{C}_i$ generated in line 03 and $\alpha_{ki}$ is the project of instantiation $\alpha$ on sepset $\mathcal{S}_{ki}$. Note here that $\alpha$ is guaranteed to be consistent with $\phi$ and $\beta$ (see the pseudocode of GENERATE_INSTANTIATIONS).

All we need to show then is that this computed expression is equivalent to $Cons^{\Delta_{ij}}(\beta \wedge \phi_{ij})$. By observing that $\Delta_{ij}$ can be decomposed into $\Delta_i$ and $\Delta_{ki}$ where $k \neq j$, this equivalence can be proven using the Intersection Corollary. Specifically, given that the atoms in clique $\mathcal{C}_i$ contain all atoms that are common between any two subtrees $\mathcal{T}_{ki}$, the Corollary gives:

$$Cons^{\Delta_{ij}}(\beta \wedge \phi_{ij}) \equiv \bigvee_\alpha Cons^{\Delta_i}(\alpha \wedge (\beta \wedge \phi_{ij})_i) \wedge \bigwedge_{k \neq j} Cons^{\Delta_{ki}}(\alpha_{ki} \wedge (\beta \wedge \phi_{ij})_{ki}).$$

Note that $(\phi_{ij})_i = \phi_i$ and $(\phi_{ij})_{ki} = \phi_{ki}$. This leads to

$$Cons^{\Delta_{ij}}(\beta \wedge \phi_{ij}) \equiv \bigvee_{\alpha} Cons^{\Delta_i}(\alpha \wedge \beta_i \wedge \phi_i) \wedge \bigwedge_{k \neq j} Cons^{\Delta_{ki}}(\alpha_{ki} \wedge \beta_{ki} \wedge \phi_{ki}).$$

The atoms in $\beta_i \wedge \phi_i$ are a subset of the atoms in $\alpha$. Therefore, $\alpha \models \beta_i \wedge \phi_i$ given that $\alpha$ and $\beta_i \wedge \phi_i$ are consistent. Similarly, the atoms in $\beta_{ki}$ are a subset of the atoms in $\alpha_{ki}$. Therefore, $\alpha_{ki} \models \beta_{ki}$ given that $\alpha_{ki}$ and $\beta_{ki}$ are consistent. This leads to

$$Cons^{\Delta_{ij}}(\beta \wedge \phi_{ij}) \equiv \bigvee_{\alpha} Cons^{\Delta_i}(\alpha) \wedge \bigwedge_{k \neq j} Cons^{\Delta_{ki}}(\alpha_{ki} \wedge \phi_{ki}). \quad \square$$

## Proof of Theorem 9

To prove that the function SYSTEM_CONSEQUENCE terminates, it is enough to prove that the calls to SYSTEM_CONSEQUENCE will not recurse infinitely. To show this, note that all recursive calls made by SUBTREE_CONSEQUENCE$(\mathcal{C}_i, \mathcal{C}_j)$ are of the form SUBTREE_CONSEQUENCE$(\mathcal{C}_x, \mathcal{C}_y)$ where the arc $\mathcal{C}_x$–$\mathcal{C}_y$ belongs to the subtree $\mathcal{T}_{ij}$. Therefore, the number of arcs visited by a recursive call is smaller than the number of arcs visited by its parent call. The boundary condition is when a call SUBTREE_CONSEQUENCE$(\mathcal{C}_i, \mathcal{C}_j)$ is made and subtree $\mathcal{T}_{ij}$ has a single clique in it.

Proving that the function SYSTEM_CONSEQUENCE returns the desired consequence follows directly from Lemma 2. Consider the jointree $\mathcal{T}'$ and its corresponding $\Delta'$ that results from adding the clique $\mathcal{C}_0 = \emptyset$ as a neighbor to the pivot clique $\mathcal{C}_z$. We have $\mathcal{T}'_{z0} = \mathcal{T}$, $\Delta'_{z0} = \Delta$, and $\phi_{z0} = \phi$. Moreover, the sepset $\mathcal{S}_{z0}$ is empty and it has one instantiation $true$. Therefore, by Lemma 2, the call on line 03 of SYSTEM_CONSEQUENCE returns $Cons^{\Delta'_{z0}}(true \wedge \phi_{z0}) = Cons^{\Delta}(\phi). \quad \square$

## Proof of Theorem 10

Suppose that each component consequence CONSEQUENCES$_P[.]$ is in decomposable NNF.

That the sentence returned by SYSTEM_CONSEQUENCE is in NNF follows immediately given that only disjunctions and conjunctions are constructed in Figure 8 (no negations).

To prove that the sentence is decomposable, we need three results:

- No assumables are shared between component descriptions, which follows from the definition of a structured system description.

- The NNF returned by CLIQUE_CONSEQUENCE$(\mathcal{C}_i)$ mentiones only assumables that appear in component descriptions assigned to clique $\mathcal{C}_i$. This follows immediately from examining the pseudocode of CLIQUE_CONSEQUENCE$(\mathcal{C}_i)$.

- The NNF returned by SUBTREE_CONSEQUENCE$(\mathcal{C}_i, \mathcal{C}_j)$ mentions only assumables that appear in component descriptions assigned to cliques in subtree $\mathcal{T}_{ij}$. This can be shown by induction on the structure of a jointree.

To prove that the returned NNF is decomposable, all we need to show is that whenever a conjunction is constructed in Figure 8, the conjuncts are guaranteed to share no assumables. There are two places where conjunctions are constructed:

1. CLIQUE_CONSEQUENCE: *Conj* is set to the conjunction of consequences that correspond to distinct component descriptions. Therefore, they are guaranteed to share no assumables.

2. SUBTREE_CONSEQUENCE: *Conj* is set to the conjunction of a NNF returned by CLIQUE_CONSEQUENCE($\mathcal{C}_i$) together with NNFs returned by SUBTREE_CONSEQUENCE($\mathcal{T}_{ki}$). These NNFs are guaranteed to share no assumables since clique $\mathcal{C}_i$ and subtrees $\mathcal{T}_{ki}$ share no assumables.

Therefore, the NNF returned by SYSTEM_CONSEQUENCE must be decomposable. □

**Lemma 4** *In Figure 8, the number of calls to* SUBTREE_CONSEQUENCE($\mathcal{C}_i, \mathcal{C}_j$) *that will not return from line 01 is no more than* $2^{|\mathcal{S}_{ij} \setminus \mathbf{E}|}$. *We will refer to such calls as* <u>non-cached calls.</u>

## Proof of Lemma 4

By Lemma 2, the sepset $\mathcal{S}_{ij}$ is instantiated when SUBTREE_CONSEQUENCE($\mathcal{C}_i, \mathcal{C}_j$) is called. Moreover, the index of such instantiation is the key used in the cache lookup on line 01 of SUBTREE_CONSEQUENCE. The total number of possible keys is $2^{|\mathcal{S}_{ij} \setminus \mathbf{E}|}$ since this is the maximum number of distinct instantiations generated for $\mathcal{S}_{ij}$ (note that atoms $\mathbf{E}$ are instantiated during the run of the algorithm). In case of a cache hit on line 01, the function returns immediately and the call is considered cached. In case of a cache miss, the call is non-cached and it will lead to an insertion into the cache on line 11 of SUBTREE_CONSEQUENCE. Since there are no more than $2^{|\mathcal{S}_{ij} \setminus \mathbf{E}|}$ keys, there are no more than $2^{|\mathcal{S}_{ij} \setminus \mathbf{E}|}$ cache insertions and, hence, no more than $2^{|\mathcal{S}_{ij} \setminus \mathbf{E}|}$ cache misses. Therefore, the maximum number of non-cached calls is $2^{|\mathcal{S}_{ij} \setminus \mathbf{E}|}$. □

## Proof of Theorem 11

We will use the following notation:

- $d_i =_{def} \sum\limits_{O \in \text{COMPONENTS\_OF}(\mathcal{C}_i)} | \mathcal{G}_O \cup \{O\} |.$

- $e_{ij} =_{def} d_i + | \mathcal{C}_i \setminus \mathcal{S}_{ij} \setminus \mathbf{E} | + \sum\limits_{k \neq j} | \mathcal{S}_{ki} |$ where $\mathcal{C}_k$ is a neighbor of clique $\mathcal{C}_i$.

Consider the following observations:

- The call GENERATE_INSTANTIATIONS($\mathcal{C}_i$) on line 03 of SUBTREE_CONSEQUENCE takes $O(2^{|\mathcal{C}_i \setminus \mathcal{S}_{ij} \setminus \mathbf{E}|})$ time, which is also the number of instantiations it returns and the number of times that lines 04-10 of SUBTREE_CONSEQUENCE will repeat.

- The calls ASSERT($\alpha$) and RETRACT($\alpha$) on lines 04 and 10 of SUBTREE_CONSEQUENCE take $O(| \mathcal{C}_i \setminus \mathcal{S}_{ij} \setminus \mathbf{E} |)$ time each since the size of $\alpha$ is $O(| \mathcal{C}_i \setminus \mathcal{S}_{ij} \setminus \mathbf{E} |)$.

- The call CLIQUE_CONSEQUENCE($\mathcal{C}_i$) on line 06 of SUBTREE_CONSEQUENCE takes $O(d_i)$ time.

- The call SUBTREE_CONSEQUENCE($\mathcal{C}_k, \mathcal{C}_i$) on line 08 takes $O(| \mathcal{S}_{ki} |)$ time <u>if it is cached.</u>

Therefore, lines 04-10 of subtree_consequence take $O(e_{ij})$ time assuming that all calls to subtree_consequence$(\mathcal{C}_k, \mathcal{C}_i)$ are cached.[24] If we assume that the number of neighbors per clique is a constant and that the number of components per clique is also a constant, then $e_{ij} = O(|\mathcal{C}_i|)$. We will indeed make this assumption in the following proof.

We first show that the time of all non-cached calls to subtree_consequence$(\mathcal{C}_i, \mathcal{C}_j)$ is

$$O(\sum_{\mathcal{C} \in \mathcal{T}_{ij}} |\mathcal{C}| \, 2^{|\mathcal{C} \setminus \mathbf{E}|}).$$

The proof is by induction on the structure of the jointree:

- *Base case:* Clique $\mathcal{C}_i$ has only one neighbor $\mathcal{C}_j$.

  Line 08 does not execute and the time of a single non-cached call to subtree_consequence$(\mathcal{C}_i, \mathcal{C}_j)$ is $O(|\mathcal{C}_i| \, 2^{|\mathcal{C}_i \setminus \mathcal{S}_{ij} \setminus \mathbf{E}|})$. By Lemma 4, there are no more than $2^{|\mathcal{S}_{ij} \setminus \mathbf{E}|}$ non-cached calls to subtree_consequence$(\mathcal{C}_i, \mathcal{C}_j)$. The total time of non-cached calls is then

  $$2^{|\mathcal{S}_{ij} \setminus \mathbf{E}|} O(|\mathcal{C}_i| \, 2^{|\mathcal{C}_i \setminus \mathcal{S}_{ij} \setminus \mathbf{E}|}) = O(|\mathcal{C}_i| \, 2^{|\mathcal{S}_{ij} \setminus \mathbf{E}| + |\mathcal{C}_i \setminus \mathcal{S}_{ij} \setminus \mathbf{E}|}),$$

  which is equal to $O(|\mathcal{C}_i| \, 2^{|\mathcal{C}_i \setminus \mathbf{E}|})$ because $\mathcal{S}_{ij}$ is a subset of $\mathcal{C}_i$.

  Since $\mathcal{C}_i$ has only one neighbor $\mathcal{C}_j$, the subtree $\mathcal{T}_{ij}$ contains only the clique $\mathcal{C}_i$. Therefore,

  $$O(|\mathcal{C}_i| \, 2^{|\mathcal{C}_i \setminus \mathbf{E}|}) = O(\sum_{\mathcal{C} \in \mathcal{T}_{ij}} |\mathcal{C}| \, 2^{|\mathcal{C} \setminus \mathbf{E}|}).$$

- *Inductive step:* Clique $\mathcal{C}_i$ has more than one neighbor.

  Suppose that for some $k$, the time of all non-cached calls to subtree_consequence$(\mathcal{C}_k, \mathcal{C}_i)$ on line 08 is

  $$O(\sum_{\mathcal{C} \in \mathcal{T}_{ki}} |\mathcal{C}| \, 2^{|\mathcal{C} \setminus \mathbf{E}|}).$$

  By Lemma 2, the time of all non-cached calls to subtree_consequence$(\mathcal{C}_i, \mathcal{C}_j)$ is then

  $$\underbrace{2^{|\mathcal{S}_{ij} \setminus \mathbf{E}|} O(|\mathcal{C}_i| \, 2^{|\mathcal{C}_i \setminus \mathcal{S}_{ij} \setminus \mathbf{E}|})}_{\text{cost assuming recursive calls are cached}} \quad + \quad \underbrace{\sum_{k \neq j} O(\sum_{\mathcal{C} \in \mathcal{T}_{ki}} |\mathcal{C}| \, 2^{|\mathcal{C} \setminus \mathbf{E}|})}_{\text{cost of non-cached recursive calls}},$$

  which reduces to

  $$O(|\mathcal{C}_i| \, 2^{|\mathcal{C}_i \setminus \mathbf{E}|}) + \sum_{k \neq j} O(\sum_{\mathcal{C} \in \mathcal{T}_{ki}} |\mathcal{C}| \, 2^{|\mathcal{C} \setminus \mathbf{E}|})$$

  since $\mathcal{S}_{ij} \subseteq \mathcal{C}_i$, and then to

  $$O(\sum_{\mathcal{C} \in \mathcal{T}_{ij}} |\mathcal{C}| \, 2^{|\mathcal{C} \setminus \mathbf{E}|})$$

  since subtree $\mathcal{T}_{ij}$ consists of clique $\mathcal{C}_j$ and the subtrees $\mathcal{T}_{ki}$ where $k \neq j$.

---

24. To get the total cost of lines 04-10, we must also add the cost of non-cached calls to subtree_consequence$(\mathcal{C}_k, \mathcal{C}_i)$.

We are now ready to bound the running time of SYSTEM_CONSEQUENCE: line 01 costs $O(1)$; line 02 costs $O(|\mathbf{E}|)$ time; and line 03 is bound by the cost of all non-cached calls to SUBTREE_CONSEQUENCE$(z, 0)$, which is

$$O(\sum_{\mathcal{C} \in \mathcal{T}'_{z0}} |\mathcal{C}| \, 2^{|\mathcal{C} \setminus \mathbf{E}|}) = O(\sum_{\mathcal{C} \in \mathcal{T}} |\mathcal{C}| \, 2^{|\mathcal{C} \setminus \mathbf{E}|})$$

since $\mathcal{T}'_{z0} = \mathcal{T}$. Therefore, the time of SYSTEM_CONSEQUENCE is $O(\sum_{\mathcal{C} \in \mathcal{T}} |\mathcal{C}| \, 2^{|\mathcal{C} \setminus \mathbf{E}|})$.

This is also the space complexity of the returned NNF since the addition of either a node or an arc to the NNF takes $O(1)$ time. $\square$

## Proof of Theorem 12

- *Reflexive:*
  For all $a \in \Sigma$, $a \oplus 0 = a$ and, hence, $a \leq_{\oplus} a$.

- *Transitive:*
  Suppose that $a \leq_{\oplus} b$ and $b \leq_{\oplus} c$. Then these exists $x$ and $y$ such that $a \oplus x = b$ and $b \oplus y = c$. Moreover, $a \oplus x \oplus y = c$ and, hence, $a \leq_{\oplus} c$.

- *Anti-symmetric:*
  Suppose that $a \leq_{\oplus} b$ and $b \leq_{\oplus} a$. Then these exists $x$ and $y$ such that $a \oplus x = b$ and $b \oplus y = a$. Moreover, $a \oplus x \oplus y = a$. This leads to $x \oplus y = 0$ and, hence, $x = 0$ and $y = 0$. Therefore, $a = b$.

- *Total:*
  For all $a$ and $b$, either $a \oplus c = b$ or $a = b \oplus c$ for some unique $c$. Hence, for all $a$ and $b$, either $a \leq_{\oplus} b$ or $b \leq_{\oplus} a$. $\square$

**Lemma 5** *The addition operation of a cost function satisfies the following properties:*

1. *$a \oplus b = a$ implies $b = 0$.*

2. *If $a <_{\oplus} b$, then $a \oplus c <_{\oplus} b \oplus c$.*

## Proof of Lemma 5

1. We have $a \oplus 0 = a$. We also have that $a \oplus b = a$ for some unique $b$. Therefore, $b = 0$.

2. If $a <_{\oplus} b$, then $a \oplus x = b$ and $x \neq 0$. We then have $a \oplus x \oplus c = b \oplus c$, which leads to $a \oplus c \leq_{\oplus} b \oplus c$. Since $a \oplus x \oplus c = b \oplus c$ for a unique $x$, and since $x \neq 0$, we have $a \oplus c \neq b \oplus c$ and, hence, $a \oplus c <_{\oplus} b \oplus c$. $\square$

**Lemma 6** *In Figure 12, if the instantiations in set $\mathbf{T}$ have the same cost $c$, then the instantiations in* EXTEND$(\mathbf{T}, \mathbf{S})$ *will have the same cost $c$.* $\square$

Moreover, we will refer to the instantiations in EXTEND$(\mathbf{T}, \mathbf{S})$ as the <u>zero extensions</u> of the instantiations in $\mathbf{T}$.

The following lemma is with respect to the algorithm in Figure 12.

**Lemma 7** *After termination of the algorithm in Figure 12, we have the following for every node $\mu$ in the NNF-graph $\tau$ and every instantiation $\alpha$ in* TERMS$(\mu)$:

1. $\alpha \models \mu$;

2. COST$(\mu) = \mathcal{F}(\alpha)$; *and*

3. $\alpha$ *is an* **S**-*instantiation where* **S** *are all the atoms appearing in the NNF-graph rooted at* $\mu$.

**Proof of Lemma 7**

The proof is by induction on the structure of the NNF-graph $\tau$.

*Base case:* $\mu$ is a leaf node.

- *Case I:* $\mu$ is a literal-node. Then COST$(\mu) = \mathcal{F}($LITERAL_OF$(\mu)) = \mathcal{F}(\{$LITERAL_OF$(\mu)\})$ and TERMS$(\mu) = \{\{$LITERAL_OF$(\mu)\}\}$. The properties hold.

- *Case II:* $\mu \equiv true$ is an and-node. Then COST$(\mu) = 0 = \mathcal{F}(\{\})$ and TERMS$(\mu) = \{\{\}\}$. The properties hold.

- *Case III:* $\mu \equiv false$ is an or-node. Then COST$(\mu) = \infty$ and TERMS$(\mu) = \{\}$. The properties hold.

*Inductive step:* $\mu$ is a node with children.

Suppose that the property holds for each child $\mu_i$ of $\mu$ and that $\alpha \in$ TERMS$(\mu)$.

- *Case I:* $\mu$ is an and-node. Then $\alpha$ has the form $\bigwedge_i \alpha_i$ where $\alpha_i \in$ TERMS$(\mu_i)$.

  1. By the induction hypothesis, $\alpha_i \models \mu_i$. Hence, $\bigwedge_i \alpha_i \models \bigwedge_i \mu_i$ and $\alpha \models \mu$ since $\mu \equiv \bigwedge_i \mu_i$.

  2. By the induction hypothesis, COST$(\mu_i) = \mathcal{F}(\alpha_i)$. From line 04 of PRUNE, COST$(\mu) = \bigoplus_i$ COST$(\mu_i)$ and, hence, COST$(\mu) = \bigoplus_i \mathcal{F}(\alpha_i) = \mathcal{F}(\alpha)$.

  3. $\mathbf{S} = \bigcup_i \mathbf{S}_i$ where $\mathbf{S}_i$ are the atoms appearing in child $\mu_i$. By the induction hypothesis, each $\alpha_i$ is an $\mathbf{S}_i$-instantiation. Therefore, $\alpha$ must be an $\mathbf{S}$-instantiation.

- *Case II:* $\mu$ is an or-node. Then $\alpha$ is the zero extension of some $\alpha_i \in$ TERMS$(\mu_i)$ where $\mu_i$ is a child of $\mu$ and COST$(\mu) =$ COST$(\mu_i)$.

  1. By the induction hypothesis, $\alpha_i \models \mu_i$. Therefore, $\alpha \models \alpha_i \models \mu_i \models \bigvee_i \mu_i \models \mu$.

  2. By the induction hypothesis, COST$(\mu_i) = \mathcal{F}(\alpha_i)$ and by Lemma 6, $\mathcal{F}(\alpha_i) = \mathcal{F}(\alpha)$. Hence, COST$(\mu_i) = \mathcal{F}(\alpha)$. Since COST$(\mu_i) =$ COST$(\mu)$, we must then have COST$(\mu) = \mathcal{F}(\alpha)$.

  3. By the induction hypothesis, each $\alpha_i$ is an $\mathbf{S}_i$-instantiation where $\mathbf{S}_i$ are the atoms appearing in $\mu_i$. By calling EXTEND(TERMS$(\mu_i), Atoms(\mu, \mu_i)$), we are extending each $\alpha_i \in$ TERMS$(\mu_i)$ with one literal for each atom in $Atoms(\mu, \mu_i)$. Since $Atoms(\mu, \mu_i) = \mathbf{S} \setminus \mathbf{S}_i$, the result of this extension must then be an $\mathbf{S}$-instantiation. $\square$

**Proof of Theorem 13**

In this theorem, *MinInst* is the set of minimal instantiations with respect to the cost function $(Atoms(\mu), \Sigma, \oplus, \mathcal{F})$ where $Atoms(\mu)$ are all atoms appearing in the NNF-graph rooted at $\mu$.

The proof is by induction on the structure of $\tau$.

*Base case:*  $\mu$ is a leaf node.

- *Case I:* $\mu$ is a literal-node. Then $\text{TERMS}(\mu) = \{\{\text{LITERAL\_OF}(\mu)\}\} = MinInst(\mu)$.

- *Case II:* $\mu \equiv true$ is an and-node. Then $\text{TERMS}(\mu) = \{\{\}\} = MinInst(\mu)$.

- *Case III:* $\mu \equiv false$ is an or-node. Then $\text{TERMS}(\mu) = \{\} = MinInst(\mu)$.

*Inductive step:*  $\mu$ is a node with children.

Suppose that $\text{TERMS}(\mu_i) = MinInst(\mu_i)$ for each child $\mu_i$ of $\mu$.

*First direction:*  If $\alpha \in MinInst(\mu)$, then $\alpha \in \text{TERMS}(\mu)$.

Suppose that $\alpha \in MinInst(\mu)$. Then $\alpha \models \mu$.

- *Case I:* $\mu$ is an and-node.
  Since $\tau$ is a decomposable NNF, the atoms that appear in each $\mu_i$ must be disjoint. Therefore, $\alpha$ must have the form $\bigwedge_i \alpha_i$ where $\alpha_i$ is the projection of $\alpha$ on the atoms in $\mu_i$. Moreover, we must have $\alpha_i \models \mu_i$ for each $\alpha_i$. It suffices then to show that $\alpha_i \in \text{TERMS}(\mu_i)$. Suppose that $\alpha_i \notin \text{TERMS}(\mu_i)$ for some $\alpha_i$. By the induction hypothesis, $\alpha_i \notin MinInst(\mu_i)$. Since $\alpha_i \models \mu_i$, we must then have $\text{COST}(\mu_i) <_\oplus \mathcal{F}(\alpha_i)$. Now let $\alpha'$ be the result of replacing this $\alpha_i$ in $\alpha$ with some $\beta_i \in \text{TERMS}(\mu_i)$. Then $\alpha' \models \mu$ since $\beta_i \models \mu_i$ by Lemma 7. Moreover, the cost of $\alpha'$ must be less than the cost of $\alpha$ by Lemma 5. Therefore, $\alpha$ cannot be in $MinInst(\mu)$ which is a contradiction. We then conclude that $\alpha_i \in \text{TERMS}(\mu_i)$ and, hence, that $\alpha \in \text{TERMS}(\mu)$.

- *Case II:* $\mu$ is an or-node.
  Since $\alpha \models \mu$, we must have $\alpha \models \mu_i$ for some child $\mu_i$ of $\mu$. Let $\alpha_i$ be the projection of $\alpha$ on the atoms in $\mu_i$. Then $\mathcal{F}(\alpha_i) \leq_\oplus \mathcal{F}(\alpha)$, $\alpha_i \models \mu_i$ and we have one of two cases:

  1. $\alpha_i \in \text{TERMS}(\mu_i)$: Then $\mathcal{F}(\alpha_i) = \text{COST}(\mu_i)$ by Lemma 7 and we have two cases:

     (a) $\mathcal{F}(\alpha_i) =_\oplus \mathcal{F}(\alpha)$: $\alpha$ is then a zero extension of $\alpha_i$ and, hence, $\alpha \in \text{TERMS}(\mu)$.

     (b) $\mathcal{F}(\alpha_i) <_\oplus \mathcal{F}(\alpha)$: any zero extension of $\alpha_i$ will both entail $\mu$ and have a lower cost than $\alpha$. This contradicts $\alpha \in MinInst(\mu)$ and the case is impossible.

  2. $\alpha_i \notin \text{TERMS}(\mu_i)$: By the induction hypothesis, $\alpha_i \notin MinInst(\mu_i)$. Since $\alpha_i \models \mu_i$, this means that $\text{COST}(\mu_i) <_\oplus \mathcal{F}(\alpha_i)$. Therefore, the zero extension of any $\beta_i$ in $\text{TERMS}(\mu_i)$ will both entail $\mu$ and have a smaller cost than $\alpha$. This contradicts $\alpha \in MinInst(\mu)$ and the case is impossible.

*Second direction:*  If $\alpha \in \text{TERMS}(\mu)$, then $\alpha \in MinInst(\mu)$.

Suppose that $\alpha \in \text{TERMS}(\mu)$. We need to show two things:

- $\alpha \models \mu$:
  This follows by Lemma 7.

- $\mathcal{F}(\alpha) = \mathcal{F}(\gamma)$ for some $\gamma \in MinInst(\mu)$:
  Suppose that $\gamma \in MinInst(\mu)$. We just showed above that $\gamma \in \text{TERMS}(\mu)$. By Lemma 7, all instantiations in $\text{TERMS}(\mu)$ have the same cost, which is also the cost of $\gamma$. Therefore, $\alpha$ and $\gamma$ must have the same cost.

This lead to $\alpha \in MinInst(\mu)$. $\square$

**Proof of Theorem 14**

The call to PRUNE on line 01 takes $O(E)$ time. Note that saving the computed costs is essential since $\tau$ is a graph, not a tree. Therefore, without the check on line 01 of PRUNE, the cost of a node may be computed more than once since a node can have more than one parent.

The call to INSTANTIATIONS on line 02 takes $O(\mid \text{TERMS}(\tau) \mid^2 E)$ time where the explanation is given below. The total time of the extraction algorithm is then $O(\mid \text{TERMS}(\tau) \mid^2 E)$.

INSTANTIATIONS is similar to PRUNE except that the amount of work done at each node $\mu$ is different. To bound this amount, we first observe that for any node $\mu$ with $cost(\mu) \neq \infty$ and its child $\nu$:[25]

$$\mid \text{TERMS}(\nu) \mid \leq \mid \text{TERMS}(\mu) \mid . \tag{1}$$

This follows because:

- if $\mu$ is an and-node, then $\text{TERMS}(\mu)$ is the cartesian product of all $\text{TERMS}(\nu)$ and, hence, its cardinality cannot be less than the cardinality of any $\text{TERMS}(\nu)$; and

- if $\mu$ is an or-node, then $\text{TERMS}(\mu)$ is the union of all $\text{EXTEND}(\text{TERMS}(\nu), Atoms(\mu, \nu))$ and, hence, its cardinality cannot be less than the cardinality of any $\text{TERMS}(\nu)$.[26]

Therefore, the computation of $\text{TERMS}(\mu)$ takes

- $O(\mid \text{TERMS}(\mu) \mid)$ time if $\mu$ is an and-node, which is also the size of the cartesian product

$$\text{TERMS}(\mu) = \bigcap_{\nu \in \text{CHILDREN}(\mu)} \text{TERMS}(\nu).$$

- $O(\mid \text{CHILDREN}(\mu) \mid \mid \text{TERMS}(\mu) \mid^2)$ time if $\mu$ is an or-node, which is justified as follows:

  - the union of two sets with sizes $n$ and $m$ takes $O(nm)$ time.

  - to compute $\text{TERMS}(\mu)$ we must perform $O(\mid \text{CHILDREN}(\mu) \mid)$ union operations.

  - $\mid \text{TERMS}(\nu) \mid \leq \mid \text{TERMS}(\mu) \mid$ for all $\nu$ in $\text{CHILDREN}(\mu)$.

---

25. If $cost(\mu) = \infty$, $\text{TERMS}(\nu) = \{\}$ and the case is handled by line 02 of INSTANTIATIONS.
26. Note that $\mid \text{TERMS}(\nu) \mid \leq \mid \text{EXTEND}(\text{TERMS}(\nu), Atoms(\mu, \nu)) \mid$.

Therefore, the time to compute TERMS$(\mu)$ is $O(|$ CHILDREN$(\mu)$ $|$ $|$ TERMS$(\mu)$ $|^2)$ in the worst case. The total time taken by INSTANTIATIONS is then

$$\sum_{\mu \in \tau} O(|\text{ CHILDREN}(\mu) \mid \mid \text{TERMS}(\mu) \mid^2),$$

where $\mu$ denotes a node in the NNF-graph $\tau$. This reduces to:

$$\sum_{\mu \in \tau} O(|\text{ CHILDREN}(\mu) \mid \mid \text{TERMS}(\tau) \mid^2)$$

since $|$ TERMS$(\mu)$ $|\leq|$ TERMS$(\tau)$ $|$ for any node $\mu$ in the NNF-graph $\tau$.[27] Reducing this further, we get:

$$|\text{ TERMS}(\tau) \mid^2 \sum_{\mu \in \tau} O(|\text{ CHILDREN}(\mu) \mid)$$
$$= \mid \text{TERMS}(\tau) \mid^2 O(E)$$
$$= O(|\text{ TERMS}(\tau)^2 \mid E). \quad \square$$

---

27. This follows by transitivity of the property in Equation 1 and because $\tau$ is an ancestor of $\mu$.